\pdfoutput=1

\PassOptionsToPackage{prologue,dvipsnames}{xcolor}
\documentclass[11pt]{article}

\usepackage[preprint]{coling}

\usepackage{times}
\usepackage{latexsym}

\usepackage[T1]{fontenc}

\usepackage[utf8]{inputenc}

\usepackage{microtype}

\usepackage{inconsolata}

\usepackage{graphicx}

\usepackage{algorithm}
\usepackage{algpseudocode}
\usepackage{amsmath}
\usepackage{svg}
\usepackage{booktabs}
\usepackage{multirow}
\usepackage{amssymb}
\usepackage{subcaption} 

\usepackage{colortbl}
\usepackage[dvipsnames]{xcolor}
\usepackage{array}
\usepackage[normalem]{ulem}
\useunder{\uline}{\ul}{}

\title{Beyond Discrete Personas: Personality Modeling Through Journal Intensive Conversations}

\author{Sayantan Pal , Souvik Das , Rohini K. Srihari \\
        State University of New York at Buffalo \\
        Department of Computer Science and Engineering \\
        \texttt{\href{mailto:spal5@buffalo.edu}{spal5}},
        \texttt{\href{mailto:souvikda@buffalo.edu}{souvikda}},
        \texttt{\href{mailto:rohini@buffalo.edu}{rohini}}@buffalo.edu}

\begin{document}
\maketitle
\begin{abstract}
Large Language Models (LLMs) have significantly improved personalized conversational capabilities. However, existing datasets like Persona Chat, Synthetic Persona Chat, and Blended Skill Talk rely on static, predefined personas. This approach often results in dialogues that fail to capture human personalities' fluid and evolving nature. To overcome these limitations, we introduce a novel dataset with around 400,000 dialogues and a framework for generating personalized conversations using long-form journal entries from Reddit. Our approach clusters journal entries for each author and filters them by selecting the most representative cluster, ensuring that the retained entries best reflect the author’s personality. We further refine the data by capturing the Big Five personality traits—openness, conscientiousness, extraversion, agreeableness, and neuroticism—ensuring that dialogues authentically reflect an individual's personality. Using Llama 3 70B, we generate high-quality, personality-rich dialogues grounded in these journal entries. Fine-tuning models on this dataset leads to an 11\% improvement in capturing personality traits on average, outperforming existing approaches in generating more coherent and personality-driven dialogues.
\end{abstract}

\section{Introduction}

\begin{figure}[htbp]
    \centering
    \includesvg[width=0.4\textwidth]{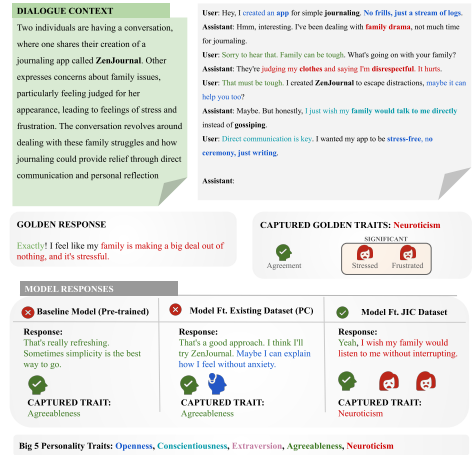}
    \caption{LLMs fine-tuned on our JIC dataset best aligns to the golden annotation capturing Personality traits compared to other models (image shows Ft. on Persona Chat). Detailed dialogue is shown in Fig. \ref{fig:main}.}
    \label{fig:intro}
\end{figure}


A conversation reflects the unique threads of a person’s life experiences, thoughts, and personalities\cite{mairesse2007using}. However, many existing conversational systems\cite{ahmad-etal-2023-rptcs, welch-etal-2020-expressive} struggle to capture the richness of these tales, often reducing complex individuals to static, predefined personas\cite{blomkvist2002persona}. Existing datasets like Persona Chat (PC)\cite{zhang-etal-2018-personalizing}, Synthetic Persona Chat (SPC)\cite{jandaghi-etal-2024-faithful-persona}, and Blended Skill Talk (BST)\cite{smith-etal-2020-put} use hardcoded personas. While these datasets have paved the way for more personalized dialogue systems\cite{kasahara-etal-2022-building}, they often fall short of capturing the dynamic and evolving nature of real human personalities\cite{allbeck2008creating}, as shown in Fig. \ref{fig:intro}. Conversations generated from such static personas can feel repetitive\cite{zhang-etal-2020-dialogpt}, shallow, and sometimes even contradictory\cite{nie-etal-2021-like}, failing to engage the user truly. Our research seeks to fill this gap and transform this approach by moving beyond the constraints of discrete personas, instead embracing a model that captures the dynamic nature of personal identity\cite{schwartz2011daily}. By leveraging long-form journal entries mined from platforms like Reddit—where individuals share their authentic, unfiltered life experiences, we ensured the preservation of personality traits, achieving greater depth and realism than static personas.


Personas are widely used to enhance\cite{li-etal-2016-persona, zhong-etal-2020-towards} user representation and conversational flow by simulating human-like dialogue. Our analysis shows that existing datasets fail to capture the complexity of the Big Five personality traits:  (O.C.E.A.N.)\cite{hurtz2000personality, azucar2018predicting}—openness, conscientiousness, extraversion, agreeableness, and neuroticism resulting in less genuine interactions.

Creating a dataset that captures personality traits is labor-intensive, traditionally relying on significant human input for persona design, conversation generation, and validation. To overcome these challenges, we utilized large language models (LLMs), specifically LLaMa 3 70B\footnote{LLaMa 3 70B is an open-source model that allows for data generation without associated costs, making it a practical choice over models like GPT-4 despite its superior performance.}\cite{llama3modelcard}, for synthetic data generation with human-in-the-loop assessment. This personalizes AI systems and enhances human-AI interaction for more relatable and engaging conversational agents(CA)\cite{clark2019makes}.


Our work introduces a novel method for creating a journal-based conversational dataset named Journal Intensive Conversations (JIC). \textbf{(1)} This process begins with data acquisition from Reddit. \textbf{(2)} We apply multi-step filtration strategies, using clustering algorithms to identify and retain the most representative journal entries per author. Additionally, we filter out dialogues that diverge significantly from the author’s average Big 5 Personality Traits, ensuring better alignment. \textbf{(3)} We then use instruct-LLMs to generate journal-grounded conversations, ensuring the resulting dialogues remain true to the author’s personality. \textbf{(4)} Finally, we demonstrate that fine-tuning state-of-the-art(SOTA) LLMs on our dataset enhances their ability to capture personality traits effectively in dialogue. Our code, data, and best models are publicly available.\footnote{\href{https://huggingface.co/choco58/LLaMAdelic}{LLaMAdelic}, \href{https://huggingface.co/choco58/MistraMystic}{MistraMystic}, \href{https://github.com/Sayantan-world/Beyond-Discrete-Personas}{Code and Data}}



\section{Related Work}

\textbf{Personality Traits in Conversational AI}: In recent years, modeling personality traits\cite{liu-etal-2016-recurrent, caron-srivastava-2023-manipulating, saha-etal-2022-stylistic} in conversational systems\cite{dusek-jurcicek-2016-context} has been an area of extensive research to make human-AI\cite{yang-etal-2024-human} interaction more personalized and engaging. Early attempts\cite{yamashita-etal-2023-realpersonachat, zhang-etal-2018-personalizing} in this field used static, predefined personas to model users and produce goal-directed faithful\cite{jandaghi-etal-2024-faithful-persona} conversations. While these systems improved personalization\cite{smith-etal-2020-put}, they fell short of capturing the dynamic nature of human behavior\cite{schill2019more, pal-etal-2024-empowering}, often reducing users to rigid attributes that limit dialogue adaptability. Recent advancements\cite{huang2024on, huang-etal-2023-learning} have called for more sophisticated systems that reflect the evolving nature of human personality. Moreover, the emergence of large language models (LLMs) like GPT-3\cite{NEURIPS2020_1457c0d6} and LLaMA\cite{touvron2023llamaopenefficientfoundation} offers new opportunities for generating more nuanced, personality-driven dialogues. Fine-tuning these models on personalized datasets enables them to exhibit a deeper understanding of individual traits, fostering more consistent and contextually appropriate interactions\cite{labruna-etal-2024-dynamic}. Other works have explored integrating psychological models \cite{azucar2018predicting, barlett2012direct} like the Big Five (O.C.E.A.N. Model) into CA. However, significant challenges remain in accurately capturing and representing dynamic personality traits. 

\textbf{Personality Datasets and Challenges}: Recent advancements in conversational datasets have highlighted the potential\cite{sun-etal-2022-safety, rashkin-etal-2019-towards} and limitations\cite{hwang-etal-2023-dialogizer} of existing approaches\cite{zhong-etal-2020-towards, das-srihari-2024-uniwiz} to simulate human behavior. Static personas or scripted inputs, such as those found in existing datasets, limit their ability to capture human traits' evolving nature in dialogue. For instance, synthetic conversations often mimic human interaction but struggle to reflect persistent personality traits over time. More dynamic datasets\cite{jandaghi-etal-2024-faithful-persona}, like those generated using instruct-LLMs, aim to address this by leveraging tunable instructions\cite{su-etal-2022-transferability} to capture authentic conversations. Despite these advancements, creating datasets that genuinely capture the complexity of human personality remains a crucial area for further research and refinement.

\section{Data Acquisition}
\begin{figure*}[htbp]
    \centering
    \includesvg[width=\textwidth]{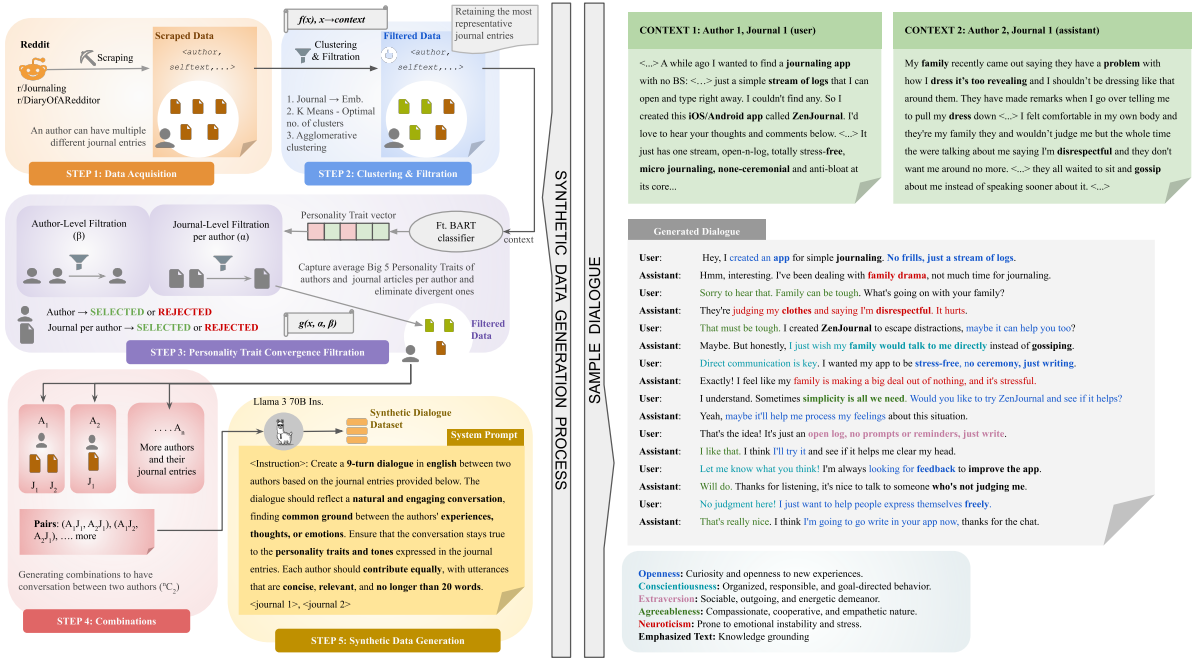}
    \caption{The synthetic data generation process is outlined in five distinct stages (left side). On the right side, we demonstrate how dialogues are generated from journal entries, highlighting the personality traits they reflect and align with. In Stage 3, where personality trait filtering is introduced, the initial values of the $\alpha$ and $\beta$ parameters were set to None to allow extensive data generation before further refinement.}
    \label{fig:main}
\end{figure*}

We mined data from two relevant subreddits: \textit{r/DiaryOfARedditor} and \textit{r/Journaling}. These communities provided a rich source of personal narratives, allowing us to gather approximately 19,000 submissions from 1,372 unique authors. 

\subsection{Journal Data Scraping}
We used the PullPush API\footnote{\href{https://pullpush.io/}{PullPush API}} to collect data from Reddit by querying relevant subreddits using a predefined keyword list. The API retrieved submissions along with details like the author's username, submission text, creation date, title, and URL. After filtering out duplicates and entries with missing fields, we obtained 18,817 submissions, including 4,377 from \textit{r/DiaryOfARedditor} and 14,440 from \textit{r/Journaling}. These submissions were further refined using additional filtering criteria.

\subsection{Synthetic Conversation Generation}



We used the Groq API's\footnote{\href{https://groq.com/}{Groq API}} LLaMa 3 70B model to generate synthetic conversations from filtered journal entries. Given rate limitations, we selected 906 out of 1,372 unique authors, pairing them in all possible combinations. For authors with multiple journal entries, dialogues were generated for every entry combination. For instance, two dialogues were generated if Author 1 had two entries and Author 2 had one. This approach produced a total of 418,476 dialogues. The final turn often included superficial exchanges like "Bye" or "Have a nice day." To retain conversational depth, the last turn was removed, leaving 8-turn dialogues that better reflected meaningful interactions. Full details of the prompting strategy are provided in Appendix \ref{apdx:prompt_synthetic}.

The synthetic dialogues were evaluated using GPT4-o and human assessments; the results showed strong agreement, particularly with high Intraclass Correlation Coefficient(ICC) scores, indicating good consistency between LLM and human ratings.  Detailed agreement scores in Appendix \ref{apdx:q_stat}.

\section{Data Filtration Strategies}

\subsection{Prominent Journal Clustering and Retention}

We employed a clustering strategy to retain the most representative journal entries for authors with multiple submissions. High-dimensional embeddings were generated using the \textit{microsoft/deberta-large} model \cite{he2021deberta} to capture semantic content. K-Means clustering, validated with silhouette scores \cite{ROUSSEEUW198753}, was applied to identify optimal clusters. Additionally, agglomerative clustering \cite{müllner2011modernhierarchicalagglomerativeclustering} was used to refine grouping, selecting the most prominent cluster. This ensured the dataset reflected each author’s dominant themes for generating synthetic conversations. The filtration process is detailed in Algorithm \ref{clustering_algo}.

\begin{algorithm}
\tiny
\caption{Prominent Journal Clustering and Retention}
\label{clustering_algo}
\begin{algorithmic}[1]
\State \textbf{Input:} Set of journal entries $\mathcal{J}$ per author, pre-trained sentence embedding model $\mathcal{M}$
\State \textbf{Initialize:} Embedding model $\mathcal{M}$ to encode journal texts into vectors $\mathbf{v}$
\For{each author $a \in \mathcal{A}$}
    \State Encode journal entries $\mathcal{J}_a$ into high-dimensional vectors $\mathbf{v}_a = \mathcal{M}(\mathcal{J}_a)$
    \State Determine the optimal number of clusters $k^*$ using K-Means clustering and silhouette score $S(k)$
    \State Apply Agglomerative Clustering with $k^*$ to group journal entries into clusters $\mathcal{C}_1, \mathcal{C}_2, \ldots, \mathcal{C}_{k^*}$
    \State Identify the largest cluster $\mathcal{C}_{\text{max}} = \arg \max_{i} |\mathcal{C}_i|$
    \State Retain journal entries from the largest cluster $\mathcal{C}_{\text{max}}$
\EndFor
\State \textbf{Output:} Subset of representative journal entries $\mathcal{C}_{\text{max}}$ for each author
\end{algorithmic}
\end{algorithm}

\subsection{Personality Trait Convergence Filtering}
We refined the dataset to capture journal entries and authors with the most prominent and consistent personality traits. Using the facebook/bart-large \cite{DBLP:journals/corr/abs-1910-13461} model, we trained a Big 5 Personality classifier with the PANDORA\cite{gjurkovic-etal-2021-pandora} dataset to evaluate and filter journal entries based on their alignment with core personality traits. The filtering process relied on two key parameters: alpha ($\alpha$), which controlled the filtration strictness at the journal level, and beta ($\beta$), which managed the convergence of personality traits across authors. The lower the parameter values, the stricter the filtration process. Algorithm \ref{convergence_algo} outlines this filtration strategy.The critical components of the process are as follows:
\begin{itemize}
    \item \textbf{Personality Trait Generation}: Each journal entry was processed using a Big 5 Personality classifier, predicting the Big 5 traits: OCEAN. This provided a detailed personality profile for each author across all their entries.
    \item \textbf{Journal-Level Filtration ($\alpha$)}: We measured each journal entry's deviation from the author's average personality profile, with $\alpha$ (can be any value but tested with 0,1) setting a threshold based on the standard deviation of these deviations. Entries with significant deviations were excluded to retain journals that best reflected the author's core traits.
    \item \textbf{Author-Level Filtration ($\beta$)}: We assessed personality consistency across authors by comparing their average profiles to the global dataset, with $\beta$ (can be any value but tested with 0,0.5) filtering out authors with excessive divergence to ensure alignment with the overall dataset.
\end{itemize}

\begin{algorithm}
\tiny
\caption{Personality Trait Convergence Filtering}
\label{convergence_algo}
\begin{algorithmic}[1]
\State \textbf{Input:} Filtered journal entries $\mathcal{J}_a$ per author $a$, pre-trained Big 5 Personality classifier $\mathcal{C}$, parameters $\alpha$, $\beta$
\State \textbf{Initialize:} Personality classifier $\mathcal{C}$ to predict Big 5 traits $\mathbf{p}_a$ for each journal
\For{each author $a \in \mathcal{A}$}
    \For{each journal entry $j \in \mathcal{J}_a$}
        \State Compute the personality traits $\mathbf{p}_j = \mathcal{C}(j)$
    \EndFor
    \State Calculate the average personality traits $\bar{\mathbf{p}}_a = \frac{1}{|\mathcal{J}_a|} \sum_{j \in \mathcal{J}_a} \mathbf{p}_j$
    \For{each journal entry $j \in \mathcal{J}_a$}
        \State Compute divergence $\Delta_j = \|\mathbf{p}_j - \bar{\mathbf{p}}_a\|$
        \If{$\Delta_j \leq \alpha$}
            \State Retain journal entry $j$
        \EndIf
    \EndFor
\EndFor
\State Calculate the global average personality traits $\bar{\mathbf{p}}_{\text{global}} = \frac{1}{|\mathcal{A}|} \sum_{a \in \mathcal{A}} \bar{\mathbf{p}}_a$
\For{each author $a \in \mathcal{A}$}
    \State Compute divergence $\Delta_a = \|\bar{\mathbf{p}}_a - \bar{\mathbf{p}}_{\text{global}}\|$
    \If{$\Delta_a \leq \beta$}
        \State Retain author $a$ and their corresponding journal entries
    \EndIf
\EndFor
\State \textbf{Output:} Refined set of authors and journal entries with consistent personality traits
\end{algorithmic}
\end{algorithm}

\section{Dataset Statistics}
JIC consists of 418,476 dialogues, 20,000 reserved for the test set \ref{apdx:test_analysis} and the rest for training, with 3,347,808 turns and 6,695,616 utterances, averaging 8 turns and 16 utterances per dialogue. Each utterance contains about 15.48 words, resulting in an average conversation length of 247.61 words. The dataset exhibits moderate topic consistency (0.5281) and an average semantic similarity of 0.3611 between consecutive utterances, highlighting its diversity and scale. Table \ref{tab:001} shows detailed comparisons with other datasets.


\begin{table}
\tiny
\centering
\begin{tabular}{lcccc}
\toprule
\textbf{Attrib} & \textbf{PC} & \textbf{SPC} & \textbf{BST} & \textbf{JIC} \\
\midrule
\textbf{\# of Conversations} & 18,878 & 10,905 & 6,808 & 418,476 \\
\textbf{Tot. \# of Turns} & 120,361 & 152,945 & 44,959 & 3,347,808 \\
\textbf{Avg. \# of Turns} & 6.38 & 14.03 & 6.60 & 8.00 \\
\textbf{Tot. \# of Utterances} & 259,600 & 310,874 & 89,918 & 6,695,616 \\
\textbf{Avg. Utt. (conv)} & 13.75 & 28.51 & 13.21 & 16.00 \\
\textbf{Avg. Words (u)} & 11.24 & 8.75 & 13.46 & 15.48 \\
\textbf{Avg. Conv. Length (w)} & 154.56 & 249.53 & 177.83 & 247.61 \\
\textbf{Longest Conv. (u)} & 49 & 117 & 28 & 16 \\
\textbf{Shortest Conv. (u)} & 11 & 6 & 4 & 16 \\
\textbf{Longest Conv. (w)} & 477 & 637 & 422 & 581 \\
\textbf{Shortest Conv. (w)} & 41 & 60 & 24 & 16 \\
\textbf{Avg. Topic Consistency (u)} & 0.50 & 0.57 & 0.55 & 0.53 \\
\textbf{Avg. Semantic Similarity (u)} & 0.31 & 0.39 & 0.36 & 0.36 \\
\bottomrule
\end{tabular}
\caption{Comparison of various datasets across several attributes. Here PC is Persona Chat, SPC is Synthetic Persona Chat, BST is Blended Skill Talk, (u) means per utterance, and (w) means per word.}
\label{tab:001}
\end{table}

\section{Experimentation}

\begin{figure}[htbp]
    \centering
    \includesvg[width=0.5\textwidth]{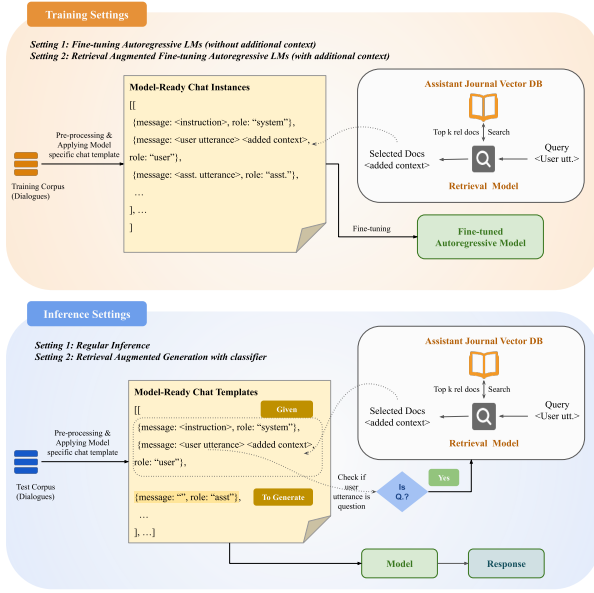}
    \caption{Model Training and Inference Settings}
    \label{fig:tr_if}
\end{figure}

Training and Inference were carried out in two settings, as shown in Fig. \ref{fig:tr_if}. The training splits for JIC are shown in Table \ref{tab:jic_splits}.

\begin{table}[htbp]
\tiny
\centering
\begin{tabular}{ccc}
\toprule
\textbf{Name} & \textbf{Abbreviation} & \textbf{Size} \\
\midrule
JIC-tiny      & JIC-t                 & 8k          \\
JIC-small     & JIC-s                 & 20k           \\
JIC-medium    & JIC-m                 & 30k           \\
JIC-large     & JIC-l                 & 100k           \\
JIC-$\alpha_{0}\beta_{0}$      & -                     & 117,749       \\
JIC-$\alpha_{1}\beta_{0}$      & JIC-best                     & 132,791       \\
JIC-$\alpha_{1}\beta_{0.5}$    & -                     & 226,873       \\
JIC-all       & -                     & 398,476   \\
\bottomrule
\end{tabular}
\caption{JIC splits and their sizes}
\label{tab:jic_splits}
\end{table}

\subsection{Training}
\label{sec:ft_jic}





To analyze the impact of data scaling, we sampled subsets of dialogues from 398,476 dialogues, consistently holding out 1,000 dialogues for validation across all experiments. Let $\mathcal{D}_\text{train}$ represent the training set and $\mathcal{D}_\text{val}$ the validation set, where $\mathcal{D}_\text{train} \subseteq \mathcal{D}$, the total dataset.
We fine-tuned the LLaMa 3 8B Instruct\cite{llama3modelcard} and Mistral 7B v0.3\cite{jiang2023mistral7b} models using a parameter-efficient technique, Low-Rank Adaptation (LoRA)\cite{hu2022lora}. Specifically, we adjusted the query, key, value, and output projection layers (i.e., $W_q$, $W_k$, $W_v$, $W_o$), updating only these parameters while keeping the rest of the model frozen.

The training objective was to minimize the negative log-likelihood (NLL) loss, defined as:
\[
\mathcal{L}(\theta) = -\frac{1}{|\mathcal{D}_\text{train}|} \sum_{(x, y) \in \mathcal{D}_\text{train}} \log p(y | x; \theta),
\]
where $x$ is the input (dialogue context) and $y$ is the output (next dialogue turn). The model aims to maximize $p(y|x)$, the probability of generating the correct response, conditioned on the context.

\textbf{LoRA Optimization:} We introduced low-rank updates to the projection layers rather than fine-tuning all weights. The update rule for $W_q$ (query projection) can be formulated as:

\[
W_q' = W_q + \Delta W_q, \quad \text{where}
\]
\[
\Delta W_q = A_q B_q, \quad A_q \in \mathbb{R}^{d \times r}, \quad B_q \in \mathbb{R}^{r \times d},
\]

with $r$ being the rank of the update. Similar updates apply to $W_k$, $W_v$, and $W_o$. This approach greatly reduces computational overhead.

We extended the training process with Retrieval Augmented Fine-tuning (RAFt.)\cite{zhang2024raftadaptinglanguagemodel} mechanisms to enhance context relevance. Let $x_i$ represent the user’s last utterance (the query) and $\mathcal{C}_i$ the assistant’s journal entry (the context). Using Maximum Marginal Relevance (MMR)\cite{article-mmr}, the top $k$ most relevant segments, denoted as $\mathcal{C}_i^{(1)}, \dots, \mathcal{C}_i^{(k)}$, were selected based on their similarity scores to $x_i$, while minimizing redundancy. The enriched input becomes:

\[
\tilde{x}_i = \text{concat}(x_i, \mathcal{C}_i^{(1)}, \dots, \mathcal{C}_i^{(k)}),
\]

where $\tilde{x}_i$ includes both the query and retrieved context. Training arguments were consistent across models and are available in Appendix \ref{apdx:tr_args}, and Training strategies for other datasets are mentioned in Appendix \ref{apdx:ft_other}.

\subsection{Inference}
Inference was carried out in two specific settings: utterance level and using Retrieval Augmented Generation(RAG) \cite{NEURIPS2020_6b493230}. RAG employed a classifier, $\mathcal{C}$, to distinguish between user statements and questions. It was activated for queries classified as questions, $\mathcal{C}(x_i) = 1$. The context $\mathcal{C}_i$ was retrieved by selecting the top $k$ relevant chunks from the assistant's journal using MMR. The enriched query becomes $\tilde{x}_i = \{x_i, \mathcal{C}_i\}$, which was passed to the model for response generation. For non-questions, $\mathcal{C}(x_i) = 0$, no retrieval was performed. This selective retrieval application improved performance, particularly in handling chit-chat vs. complex queries. 

\subsection{Evaluation Strategy}
We employed both automatic metric-based evaluation and evaluation by the LM Eval Harness\cite{eval-harness} framework by EleutherAI. Automatic evaluation used BLEU\cite{papineni-etal-2002-bleu}, METEOR\cite{banerjee-lavie-2005-meteor}, BERTScore\cite{Zhang*2020BERTScore:}, and ROUGE-1, ROUGE-2, and ROUGE-L\cite{lin-2004-rouge}. The average of these metrics provided an overall performance score. Whereas, the LM Eval Harness assessed the models on the Big 5 personality traits\footnote{Metrics: persona\_openness, persona\_conscientiousness, persona\_extraversion, persona\_agreeableness, persona\_neuroticism} across 1,000 samples per trait, highlighting the model’s ability to adapt to distinct personality profiles.

\section{Results and Discussion}


We conducted extensive testing on automated metrics and LM-eval benchmarks to assess model performance across different configurations. 

\subsection{Automatic metric-based evaluation}

\begin{table}[htbp]
\tiny
\centering
\begin{tabular}{cccccc}
\toprule
\textbf{Dataset} &
  \textbf{Model} &
  \textbf{Train cfg.} &
  \textbf{Test cfg.} &
  \textbf{Avg. Score} \\
\midrule
\multirow{2}{*}{Pre-trained} &
  LLaMA &
  ZS &
  RAG &
  0.2516 \\
\cmidrule(lr){2-5}
 &
  Mistral &
  ZS &
  Reg. &
  0.2154 \\
\midrule
\multirow{2}{*}{PC} &
  LLaMA &
  PAFt. &
  RAG &
  0.2538 \\
\cmidrule(lr){2-5}
 &
  Mistral &
  PAFt. &
  RAG &
  0.2122 \\
\midrule
\multirow{2}{*}{SPC} &
  LLaMA &
  PAFt. &
  RAG &
  0.2544 \\
\cmidrule(lr){2-5}
 &
  Mistral &
  PAFt. &
  RAG &
  0.2132 \\
\midrule
\multirow{2}{*}{BST} &
  LLaMA &
  PAFt. &
  RAG &
  0.2518 \\
\cmidrule(lr){2-5}
 &
  Mistral &
  PAFt. &
  RAG &
  0.2134 \\
\midrule
\multirow{2}{*}{JIC-$\alpha_{1}\beta_{0}$} &
  LLaMA &
  RAFt. &
  RAG &
  0.2843 \\
\cmidrule(lr){2-5}
 &
  Mistral &
  Ft. &
  RAG &
  0.2453 \\
\midrule
\multirow{2}{*}{JIC-all} &
  LLaMA &
  Ft. &
  RAG &
  \underline{\textbf{0.3105}} \\
\cmidrule(lr){2-5}
 &
  Mistral &
  Ft. &
  RAG &
  \textbf{0.2646} \\
\bottomrule
\end{tabular}
\caption{Evaluation of LLaMA and Mistral models trained on various Datasets, tested on JIC(2k subset). The highest average score across models is highlighted. The best score in the table is underlined. Only reported the best score for each configuration, detailed results for JIC in Table \ref{tab:basic_JIC_detailed}, other Datasets Table \ref{tab:basic_other_detailed}.}
\label{tab:basic_JIC}
\end{table}

\begin{figure}[htbp]
    \centering
    \resizebox{0.5\textwidth}{!}{\includesvg{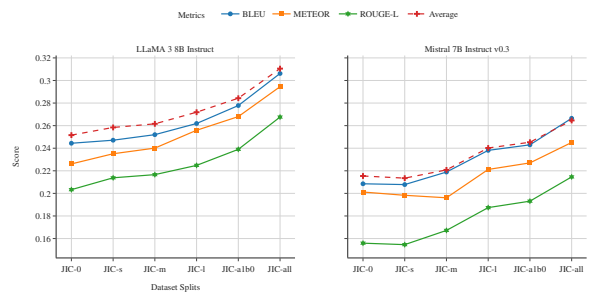}}
    \caption{Performance of LLaMA(left) and Mistral(right) models across various JIC dataset splits. Reported: BLEU, METEOR, ROUGE-L, Avg(across all: Table \ref{tab:basic_JIC}).}
    \label{fig:automated}
\end{figure}


We evaluated the LLaMA and Mistral models across different configurations and dataset sizes, with consistent validation (randomly sampled) and test splits of 1k and 2k samples. As expected, model performance generally improved as training data increased. The best average scores were achieved when models were trained on the entire dataset, with LLaMA consistently outperforming Mistral. LLaMA achieved a best average score of 0.3105, representing a 35.1\% improvement over its zero-shot baseline, while Mistral reached a best score of 0.2646, improving by 23.2\%. Table \ref{tab:basic_JIC} report our findings. Fig. \ref{fig:automated} shows comparative results, detailed results in Appendix \ref{apdx:automated_detailed}. Due to computational constraints, we could not use RAFt on the entire dataset.

Contrary to expectations, where RAFt should theoretically enhance model performance by providing additional context, Mistral did not show any improvements when using RAFt., regardless of the dataset size. This contrasts with LLaMA, which consistently improved with RAFt. across all data splits. One possible explanation for this discrepancy is that Mistral, starting from a weaker baseline, might not have been able to effectively leverage the additional retrieved context provided by the retrieval mechanism. In contrast, LLaMA’s stronger baseline performance allowed it to utilize the retrieved information more effectively, leading to consistent gains with RAFt. This suggests that while RAFt is generally beneficial, its effectiveness may depend on the model’s inherent capabilities and baseline performance.

\subsection{LM-Eval Harness Results}
\begin{table*}[htbp]
\tiny
\centering
\begin{tabular}{cccccccccc}
\toprule
\multirow{2}{*}{\textbf{Dataset}} &
  \multirow{2}{*}{\textbf{Model}} &
  \multirow{2}{*}{\textbf{Train. cfg.}} &
  \multirow{2}{*}{\textbf{Train size}} &
  \multicolumn{5}{c}{\textbf{Personality Traits Score}} &
  \multirow{2}{*}{\textbf{AVG}} \\
  \cmidrule(lr){5-9}
 &
   &
   &
   &
  \textbf{O} &
  \textbf{C} &
  \textbf{E} &
  \textbf{A} &
  \textbf{N} &
   \\
   \midrule
\multirow{2}{*}{Pre-Trained} &
  LLaMA &
  ZS &
  - &
  0.8760 &
  0.7620 &
  0.7170 &
  0.9500 &
  0.5220 &
  0.7654 \\
  \cmidrule(lr){2-10}
 &
  Mistral &
  ZS &
  - &
  0.8360 &
  0.6390 &
  0.5140 &
  0.8160 &
  0.5350 &
  0.6680 \\
  \midrule
\multirow{2}{*}{PC} &
  LLaMA &
  Ft. &
  Full &
  0.8740     
  & 0.7660     
  & 0.7180     
  & 0.9510     
  & 0.5240     
  & 0.7666 \\
  \cmidrule(lr){2-10}
 &
  Mistral &
  Ft. &
  Full &
  0.8380     
  & 0.6380     
  & 0.5140     
  & 0.8080     
  & 0.5370     
  & 0.6670 \\
  \midrule
\multirow{2}{*}{SPC} &
  LLaMA &
  PAFt. &
  Full &
  0.8750     & 
  0.7680     & 
  0.7190     & 
  0.9500     & 
  0.5240     & 
  0.7672 \\
  \cmidrule(lr){2-10}
 &
  Mistral &
  PAFt. &
  Full &
  0.8320     & 
  0.6330     & 
  0.5120     & 
  0.8100     & 
  0.5360     & 
  0.6646 \\
  \midrule
\multirow{2}{*}{BST} &
  LLaMA &
  PAFt. &
  Full &
  0.8760     & 
  0.7630     & 
  0.7160     & 
  0.9510     & 
  0.5240     & 
  0.7660 \\
  \cmidrule(lr){2-10}
 &
  Mistral &
  Ft. &
  Full &
  0.8360     & 
  0.6400     & 
  0.5130     & 
  0.8180     & 
  0.5340     & 
  0.6682 \\
  \midrule
  \multirow{4}{*}{JIC-medium} &
  \multirow{2}{*}{LLaMA} &
  Ft. &
  29k &
  0.8770 &
  0.7800 &
  0.7160 &
  0.9520 &
  0.5300 &
  0.7710 \\
 &
   &
  RAFt. &
  29k &
  0.8980 &
  \textbf{0.8040} &
  {\ul \textbf{0.7400}} &
  0.9600 &
  0.5350 &
  0.7874 \\
  \cmidrule(lr){2-10}
 &
  \multirow{2}{*}{Mistral} &
  Ft. &
  29k &
  0.8670 &
  0.6820 &
  0.5220 &
  0.8600 &
  0.5380 &
  0.6938 \\
 &
   &
  RAFt. &
  29k &
  0.9070 &
  0.7330 &
  0.5790 &
  0.8940 &
  0.5450 &
  0.7316 \\
  \midrule
\multirow{8}{*}{JIC ($\alpha$,$\beta$)} &
  \multirow{4}{*}{LLaMA} &
  Ft. $\alpha_{0}\beta_{0}$&
  $\sim$115k &
  0.8810 &
  0.7980 &
  0.7140 &
  0.9580 &
  0.5380 &
  0.7778 \\
 &
   &
  Ft. $\alpha_{1}\beta_{0}$ &
  $\sim$135k &
  0.8860 &
  0.7930 &
  0.7080 &
  {\ul \textbf{0.9600}} &
  0.5570 &
  0.7808 \\
 &
   &
  \textbf{RAFt. $\alpha_{1}\beta_{0}$*} &
  $\sim$135k &
  \textbf{0.9150} &
  0.7840 &
  0.6680 &
  0.9440 &
  {\ul \textbf{0.7040}} &
  {\ul \textbf{0.8030}} \\
 &
   &
  Ft. $\alpha_{1}\beta_{0.5}$ &
  $\sim$220k &
  0.8830 &
  0.7990 &
  0.7080 &
  0.9580 &
  0.5380 &
  0.7772 \\
  \cmidrule(lr){2-10}
 &
  \multirow{4}{*}{Mistral} &
  Ft. $\alpha_{0}\beta_{0}$ &
  $\sim$100k &
  0.9090 &
  0.7430 &
  0.5690 &
  0.9030 &
  0.5490 &
  0.7346 \\
 &
   &
  Ft. $\alpha_{1}\beta_{0}$ &
  $\sim$135k &
  0.9120 &
  0.7480 &
  0.5800 &
  0.9090 &
  0.5490 &
  0.7396 \\
 &
   &
  \textbf{RAFt. $\alpha_{1}\beta_{0}$*} &
  $\sim$135k &
  {\ul \textbf{0.9340}} &
  {\ul \textbf{0.8260}} &
  \textbf{0.6250} &
  \textbf{0.9530} &
  \textbf{0.5700} &
  \textbf{0.7816} \\
 &
   &
  Ft. $\alpha_{1}\beta_{0.5}$ &
  $\sim$220k &
  0.9050 &
  0.7530 &
  0.5760 &
  0.9140 &
  0.5590 &
  0.7414 \\
  \midrule
\multirow{2}{*}{JIC-all} &
  LLaMA &
  Ft. &
  $\sim$400k &
  0.8340 &
  0.6990 &
  0.6000 &
  0.9260 &
  0.6450 &
  0.7408 \\
  \cmidrule(lr){2-10}
 &
  Mistral &
  Ft. &
  $\sim$400k &
  0.8840 &
  0.7110 &
  0.5310 &
  0.9110 &
  0.5650 &
  0.7204 \\
\bottomrule
\end{tabular}
\caption{Scores for Big Five traits (O, C, E, A, N) are shown, with the top scores for each model highlighted and the best overall in the Table (per trait) underlined. * denotes best model. Detailed results of all JIC subsets in Table \ref{tab:lm_eval_main_detail}.}
\label{tab:lm_eval_main}
\end{table*}

\begin{figure}[htbp]
    \centering
    \resizebox{0.5\textwidth}{!}{\includesvg{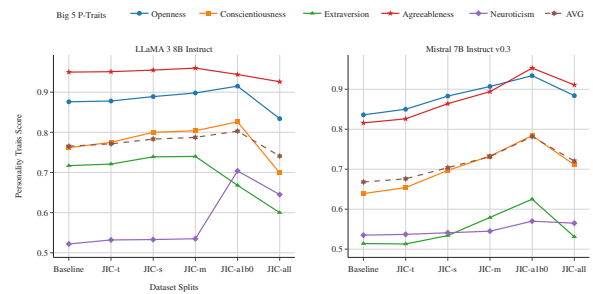}}
    \caption{Performance of LLaMA and Mistral models across various JIC dataset splits. The left panel displays the results for LLaMA, while the right panel shows the results for Mistral.}
    \label{fig:persona}
\end{figure}

\begin{figure}[htbp]
    \centering
    \resizebox{0.5\textwidth}{!}{\includesvg{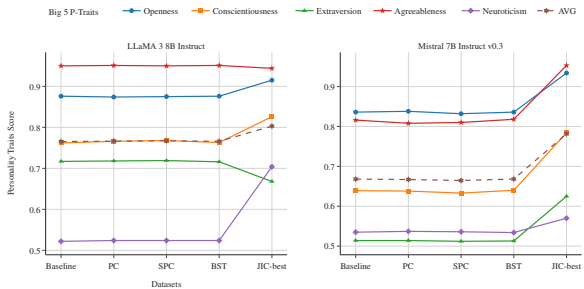}}
    \caption{Personality trait scores across various datasets for the LLaMA 3 8B Instruct model (left) and Mistral 7B Instruct v0.3 (right)}
    \label{fig:persona_oth}
\end{figure}

We analyzed the models' ability to capture Big Five personality traits (O, C, E, A, N) across different training configurations. Notably, trait scores plateaued after a certain dataset size, fine-tuned using $\alpha$ and $\beta$ parameters. The highest scores were achieved with $\alpha$=1 and $\beta$=0 (around one-third of the dataset). LLaMA outperformed Mistral, achieving a top score of 0.8030 with RAFt., while Mistral's best score was 0.7816. LLaMA's strong baseline (0.7654) showed modest improvement (4.9\%), whereas Mistral's lower baseline (0.6680) saw a 17\% gain, suggesting that Mistral benefits more from data scaling and refinement, despite LLaMA's better overall performance.These results suggest that LLaMA's strong baseline may limit its potential for further improvement, while Mistral benefits more from data scaling and refinement strategies. Table \ref{tab:lm_eval_main} report our findings. Fig. \ref{fig:persona} shows performance of LLaMA and Mistral models across various JIC splits. Fig. \ref{fig:persona_oth} compares performance of the models across different datasets.


Interestingly, while Mistral showed no improvement with RAFt in automated metrics, both models displayed significant gains in capturing personality traits with RAFt. This suggests that RAFt enables models to learn and internalize personality traits, even if the generated text doesn't exactly match the golden annotations. The retrieval process helps models better understand and generalize trait-specific behaviors, emphasizing these traits during training, regardless of text alignment.


\subsection{Ablation Study}
\begin{table*}[]
\tiny
\centering
\begin{tabular}{clcccccc}
\toprule
\multirow{2}{*}{\textbf{Model}} &
  \multicolumn{1}{c}{\multirow{2}{*}{\textbf{Variants}}} &
  \multicolumn{5}{c}{\textbf{Personality Traits Score}} &
  \multirow{2}{*}{\textbf{AVG}} \\
  \cmidrule(lr){3-7}
 & \multicolumn{1}{c}{}              & \textbf{O} & \textbf{C}      & \textbf{E}      & \textbf{A}      & \textbf{N} &        \\
 \midrule
\multirow{5}{*}{LLaMA} &
  \textbf{(0) RAFt. ($\alpha$=1,$\beta$=0)*} &
  \textbf{0.9150} &
  0.7840 &
  0.6680 &
  0.9440 &
  \textbf{0.7040} &
  {\ul \textbf{0.8030}} \\
 & (1) (w/o RA) Ft. ($\alpha$=1,$\beta$=0)        & 0.8860     & \textbf{0.7930} & 0.7080          & \textbf{0.9600} & 0.5570     & 0.7808 \\
 & (1) + (2) w/o (a,b)               & 0.8340     & 0.6990          & 0.6000          & 0.9260          & 0.6450     & 0.7408 \\
 & (1)+ (2) + (3) w/ random sampling & 0.8840     & \textbf{0.7930} & \textbf{0.7180} & 0.9550          & 0.5320     & 0.7764 \\
 & (1) + (2) + (4) w/o Ft.           & 0.8760     & 0.7620          & 0.7170          & 0.9500          & 0.5220     & 0.7654 \\
 \midrule
\multirow{5}{*}{Mistral} &
  \textbf{(0) RAFt. ($\alpha$=1,$\beta$=0)*} &
  \textbf{0.9340} &
  \textbf{0.8260} &
  \textbf{0.6250} &
  \textbf{0.9530} &
  \textbf{0.5700} &
  \textbf{0.7816} \\
 & (1) (w/o RA) Ft. ($\alpha$=1,$\beta$=0)        & 0.9120     & 0.7480          & 0.5800          & 0.9090          & 0.5490     & 0.7396 \\
 & (1) + (2) w/o (a,b)               & 0.8840     & 0.7110          & 0.5310          & 0.9110          & 0.5650     & 0.7204 \\
 & (1)+ (2) + (3) w/ random sampling & 0.905      & 0.736           & 0.563           & 0.901           & 0.546      & 0.7302 \\
 & (1) + (2) + (4) w/o Ft.           & 0.8360     & 0.6390          & 0.5140          & 0.8160          & 0.5350     & 0.6680 \\
 \bottomrule
\end{tabular}
\caption{Ablation study on JIC dataset comparing LLaMA and Mistral models across configurations, with best RAFt. settings ($\alpha$=1, $\beta$=0) highlighted and bolded top scores per trait per model.}
\label{tab:ablation}
\end{table*}

The ablation study focuses solely on personality traits, as the dataset is specifically designed to capture human-like personality dynamics, making automated metric evaluations less relevant in this context. Table \ref{tab:ablation} shows the impact of various configurations on the overall performance of LLaMA and Mistral. \textbf{(0)}: The best configuration, \textbf{RAFt.} ($\alpha=1$, $\beta=0$), yielded the highest average scores for both LLaMA (\textbf{0.8030}) and Mistral (\textbf{0.7816}). \textbf{(1)}: \textbf{No Retriever Augmentation}, the performance slightly dropped for both models (LLaMA: 0.7808, Mistral: 0.7396). \textbf{(2)}:  \textbf{No filtration} ($\alpha$, $\beta$ set to None), we observed further performance degradation (LLaMA: 0.7408, Mistral: 0.7204). \textbf{(3)}:  \textbf{Random sampling} to mimic (1) improved the scores compared to (2) but did not outperform (1), which shows the requirement of filtration (LLaMA: 0.7764, Mistral: 0.7302). \textbf{(4)}:   \textbf{No Fine-tuning}, resulted in the lowest scores (LLaMA: 0.7654, Mistral: \textbf{0.6680}).

\subsection{Personality Trait Scaling and Dataset Bias}
We found that personality trait accuracy does not scale linearly with dataset size, likely due to dataset bias. The synthetic dialogues, often reflecting negative experiences from Reddit journal entries, led to an over-representation of neuroticism in the JIC dataset, skewing the capture of other traits like extraversion and conscientiousness. LLaMA, when fine-tuned without alpha-beta filtering, showed higher neuroticism, disrupting trait balance, with Mistral exhibiting similar but less pronounced behavior. LLaMA had the highest conscientiousness and extraversion scores in the JIC-medium split, potentially due to its stronger zero-shot baseline, while Mistral required more fine-tuning for similar performance. Furthermore, focusing on personality traits slightly reduced the model's ability in general reasoning tasks, which we believe can be improved through rehearsal learning\cite{wang-etal-2023-rehearsal}. Detailed scores are in Appendix \ref{apdx:lm_eval_other_bench}. The relationship between data scaling and trait capture appears model-dependent and influenced by pre-training performance, with alpha-beta hyper-parameters ($\alpha$=1, $\beta$=0) offering the most balanced results across traits. Due to the computational costs, we experimented with a small range of values, finding the optimal remains an open research challenge.

\subsection{Qualitative \& Toxicity Analysis}

\begin{table}[]
\tiny
\centering
\begin{tabular}{>{\centering\arraybackslash}p{0.15\linewidth}|p{0.10\linewidth}p{0.4\linewidth}p{0.15\linewidth}}
\toprule
\textbf{Settings} &
  \multicolumn{1}{c}{\textbf{Role}} &
  \multicolumn{1}{c}{\textbf{Prompt \& Response}} &
  \textbf{Captured Traits} \\
  \midrule
Golden &
  Assistant &
  Exactly! I feel like my \textcolor{BrickRed}{family is making a big deal out of nothing, and it's stressful}. &
  \textcolor{BrickRed}{Neuroticism} \\
  \midrule
\textbf{LLaMA (our best)} &
  Assistant &
  Yeah, I wish my \textcolor{BrickRed}{family would listen to me without interrupting}. &
  \textcolor{BrickRed}{Neuroticism} \\
  \midrule
  LLaMA PAFt. SPC &
  Assistant &
  \textcolor{ForestGreen}{That's really refreshing. Sometimes simplicity is the best way to get things off your chest.}. &
  \textcolor{ForestGreen}{Agreeableness} \\
  \midrule
LLaMA (base) &
  Assistant &
  \textcolor{ForestGreen}{That's really refreshing. Sometimes simplicity is the best way to go}. &
  \textcolor{ForestGreen}{Agreeableness} \\
  \midrule
\textbf{Mistral (our best)} & Assistant & That's smart. I've been so \textcolor{BrickRed}{busy with family conflict}, I haven't had time to reflect.                 & \textcolor{BrickRed}{Neuroticism}             \\
\midrule
Mistral (base) \& PAFt. SPC              & Assistant & That's a \textcolor{ForestGreen}{good approach}. \textcolor{RoyalBlue}{I think I'll try ZenJournal}. Maybe I can explain how I feel without anxiety. & \textcolor{ForestGreen}{Agreeableness}, \textcolor{RoyalBlue}{Openness} \\
\bottomrule
\end{tabular}
\caption{Comparison of real and model-generated dialogues capturing personality traits. The Table demonstrates how our best-performing models (LLaMA and Mistral) align with the traits reflected in the original dialogue. Detailed in  Table \ref{tab:q001}.}
\label{tab:qual}
\end{table}

Qualitative analysis is crucial for evaluating LLMs beyond quantitative metrics, offering insights into dialogue subtleties. Table \ref{tab:qual} shows how well the models aligned with personality traits from a randomly selected annotated dialogue, ensuring unbiased and representative results. Examples can be found in Appendix \ref{apdx:q_eg}. Additionally, we conducted a toxicity analysis using the Detoxify\footnote{\href{https://github.com/unitaryai/detoxify}{Detoxify}} library to assess harmful content in our dataset, classifying dialogues based on various categories such as toxic, severely toxic, obscene, insult, identity hate, and threat. Dialogues with more than 25\% toxic utterances were flagged, and those with severe issues like threats or identity hate were also marked. We found around 11k utterances out of 6.7M utterances classified as toxic. Flagged dialogues will be kept separately during release (detailed results in Appendix \ref{apdx:toxic}).

\section{Conclusion}
Our research introduces the JIC dataset, which overcomes the limitations of static personas in existing conversational datasets. By grounding dialogues in long-form journal entries and capturing dynamic personality traits through a multi-step filtering process, we enable LMs to generate more authentic, personalized conversations. This approach significantly enhances conversational AI's ability to reflect real human personalities, offering engaging and relatable interactions.

\section*{Limitations}
\begin{itemize}
    \item A fundamental limitation of our research lies in tuning the optimal $\alpha$ and $\beta$ parameters. While the chosen values ($\alpha=1$, $\beta=0$) yielded promising results, refining these parameters remains an open challenge due to the computational demands of extensive experimentation.
    \item Synthetic data generation using LLaMA 70B introduces potential biases and safety concerns inherent in the pre-trained model. These biases could propagate into the dialogues, limiting diversity and authenticity, although the impact may be minimal.
    \item Furthermore, the dataset’s inherent bias—stemming from an over-representation of neuroticism in Reddit journals—may have skewed the models’ ability to capture traits like extraversion accurately.
    \item Finally, human evaluation remains a significant challenge, as assessing nuanced traits in synthetic dialogues can be tough and labor-intensive, highlighting the difficulty of balancing human insight and scalable evaluation methods.
\end{itemize}

\section*{Ethical Considerations}
In creating the JIC dataset, we ensured that all journal entries were publicly available and anonymized to protect user privacy. We employed the Detoxify library to tag potentially toxic dialogues to mitigate the risk of harmful content. A strict threshold was set, flagging any dialogue where more than 25\% of the utterances were classified as toxic. These flagged dialogues were kept separately to prevent their use in downstream tasks. This approach helps ensure the dataset remains safe and responsible for use in developing conversational AI systems.

\section*{Acknowledgements}
We thank the anonymous reviewers for providing valuable feedback on our manuscript. We also extend our gratitude to the human judges who contributed to the evaluation of LLM-generated synthetic data. The human evaluation team was led by Sutirtho Roy (PhD student, Literatures in English, UC San Diego) and included Anushka Tiwari (PhD student, Computational and Data-Enabled Sciences, University at Buffalo). This work is supported by NSF Grant \# 2214070. The content in this paper is solely the responsibility of the authors and does not necessarily represent the official views of the funding entity.

\bibliography{anthology_0, anthology_1, custom}

\appendix



\section{Prompting Strategy }
\label{apdx:prompt_synthetic}
The prompting strategy was carefully designed to guide the instruct models in generating dialogues that align with the personality traits and tones expressed in the journal entries. By explicitly framing the instruction to focus on finding common ground between the authors’ experiences, thoughts, or emotions, we ensured that the models would remain faithful to the context provided by the journals. This was crucial for maintaining the natural flow and personality-driven nature of the conversation. The instruction emphasized creating a balanced, engaging dialogue that reflected the distinct personality traits evident in the journals. This helped direct the model toward producing conversations that stayed true to the underlying personalities, encouraging the generation of responses that aligned with each author's emotional tone and life experiences. \textbf{Note}: Due to the limitations of the free API, some dialogues were incomplete or incorrect, leading us to skip certain combinations. As a result, the actual number of dialogues generated is lower than the possible maximum combinations.

The detailed prompt provided is as follows:
\begin{quote}
<Instruction>: Create a 9-turn dialogue in english between two authors based on the journal entries provided below. The dialogue should reflect a natural and engaging conversation, finding common ground between the authors' experiences, thoughts, or emotions. Ensure that the conversation stays true to the personality traits and tones expressed in the journal entries. Each author should contribute equally, with utterances that are concise, relevant, and no longer than 20 words.
<journal 1>, <journal 2>
\end{quote}


\section{Quality Assessment of Synthetic Dialogues \& Statistical Significance}
\label{apdx:q_stat}

\begin{figure}[htbp]
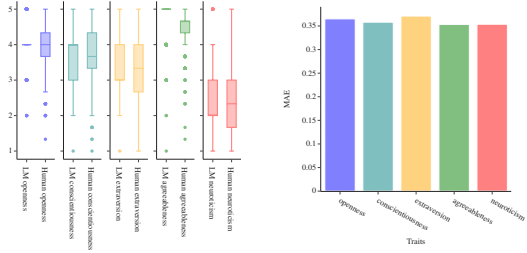

    \centering
    \begin{subfigure}[b]{0.23\textwidth}
        \centering
        \resizebox{\textwidth}{!}{\includesvg{images/boxplot.svg}}
    \end{subfigure}
    \begin{subfigure}[b]{0.23\textwidth}
        \centering
        \resizebox{\textwidth}{!}{\includesvg{images/mae.svg}}
    \end{subfigure}
    \caption{Left: Boxplot comparison of personality trait ratings (OCEAN) between the LLM (GPT-4o) and human annotators. Right: Mean Absolute Error (MAE) between LLM and Human Scores for Big 5 Traits}
    \label{fig:box_mae}
\end{figure}


We used an LLM-based evaluation of 1,000 randomly sampled synthetic dialogues from our dataset using a Likert scale(1-5), focusing on O.C.E.A.N criteria. Three in-house PhD students from linguistics and computer science rated these dialogues. 
The box plot in Fig. \ref{fig:box_mae} shows that human rating had higher variability than LLM rating. The MAE plot highlights almost similar errors for all the traits. Table \ref{tab:quality_eg} shows one example dialogue (not cherry-picked) rated by LLM and Human judges. 

Furthermore, we assessed 4000 more randomly sampled dialogues using GPT-4o. Table \ref{tab:mean_std} shows the Mean and Standard Deviation across various traits. We evaluated the agreement between the language model's (LM) personality trait predictions and human annotations using Pearson, Spearman, and Intraclass Correlation Coefficients (ICC). As shown in Table \ref{tab:icc}, Neuroticism had the highest LM-human agreement with an ICC of 0.970. These results demonstrate that the LM effectively captures most personality traits, though certain traits, like Openness and Agreeableness, exhibit slightly lower alignment with human annotations.

\begin{table*}[]
\tiny
\centering
\begin{tabular}{ccccccccccccc}
\toprule
\multirow{2}{*}{\textbf{Setting}} & \multirow{2}{*}{\textbf{Metric}} & \multicolumn{6}{c}{\textbf{Basic}}                  & \multicolumn{5}{c}{\textbf{Personality}}   \\
\cmidrule(lr){3-8} \cmidrule(lr){9-13}
 &
   &
  \textbf{Coh} &
  \textbf{Grnd} &
  \textbf{Spc} &
  \textbf{Flu} &
  \textbf{Und} &
  \textbf{Eng} &
  \textbf{O} &
  \textbf{C} &
  \textbf{E} &
  \textbf{A} &
  \textbf{N} \\
  \midrule
\multirow{2}{*}{LLM ($\sim$5k)}   & Mean                             & 4.9452 & 4.9532 & 4.2390 & 4.9898 & 4.9928 & 4.1592 & 4.0296 & 3.7690 & 3.2711 & 4.8012 & 2.4039 \\
                                  & Std                              & 0.2534 & 0.2232 & 0.4715 & 0.1005 & 0.0846 & 0.4289 & 0.4996 & 0.6451 & 0.5766 & 0.4571 & 1.0225 \\
                                  \midrule
\multirow{2}{*}{LLM (1k)}         & Mean                             & 4.9369 & 4.9409 & 4.2352 & 4.9860 & 4.9920 & 4.1722 & 4.0611 & 3.7738 & 3.3213 & 4.8228 & 2.3493 \\
                                  & Std                              & 0.2884 & 0.2714 & 0.4628 & 0.1258 & 0.0998 & 0.4345 & 0.5175 & 0.6354 & 0.5870 & 0.4405 & 1.0542 \\
                                  \midrule
\multirow{2}{*}{Humans (1k)}      & Mean                             & -      & -      & -      & -      & -      & -      & 3.9957 & 3.7287 & 3.3146 & 4.5162 & 2.3974 \\
                                  & Std                              & -      & -      & -      & -      & -      & -      & 0.6209 & 0.7281 & 0.7695 & 0.4466 & 1.0548 \\
                                  \bottomrule
\end{tabular}
\caption{Mean and Standard Deviation across various traits. Here Coherence, Grounding, Specificity, Fluency, Understandability and Engagement are abbreviated as Coh, Grnd, Spc, Flu, Und and Eng respectively}
\label{tab:mean_std}
\end{table*}

\begin{table*}[]
\tiny
\centering
\begin{tabular}{l|ccccc}
\toprule
 \textbf{Configs} &
  \textbf{Openness} &
  \textbf{Conscientiousness} &
  \textbf{Extraversion} &
  \textbf{Agreeableness} &
  \textbf{Neuroticism} \\ 
  \midrule
\textbf{P\_LLM\_J1}        & 0.7711  & 0.7933  & 0.7916  & 0.7684  & 0.9189  \\
\textbf{S\_LLM\_J1}        & 0.7638  & 0.7823  & 0.8023  & 0.7034  & 0.9142  \\
\textbf{P\_LLM\_J2}        & 0.7351  & 0.8174  & 0.7982  & 0.7702  & 0.9280  \\
\textbf{S\_LLM\_J2}        & 0.7241  & 0.8135  & 0.8172  & 0.7084  & 0.9218  \\
\textbf{P\_LLM\_J3}        & 0.7475  & 0.8287  & 0.8005  & 0.7946  & 0.9295  \\
\textbf{S\_LLM\_J3}        & 0.7395  & 0.8209  & 0.8202  & 0.7348  & 0.9243  \\
\textbf{P\_J1\_J2}         & 0.5763  & 0.6521  & 0.6373  & 0.5745  & 0.8487  \\
\textbf{S\_J1\_J2}         & 0.5593  & 0.6437  & 0.6628  & 0.4897  & 0.8353  \\
\textbf{P\_J1\_J3}         & 0.5660  & 0.6508  & 0.6282  & 0.6289  & 0.8593  \\
\textbf{S\_J1\_J3}         & 0.5600  & 0.6361  & 0.6588  & 0.5279  & 0.8499  \\
\textbf{P\_J2\_J3}         & 0.5349  & 0.6592  & 0.6354  & 0.5971  & 0.8699  \\
\textbf{S\_J2\_J3}         & 0.5200  & 0.6538  & 0.6643  & 0.5039  & 0.8602  \\
\textbf{ICC}               & 0.8751  & 0.9124  & 0.9020  & 0.8914  & 0.9703  \\
\bottomrule
\end{tabular}
\caption{Pearson(P), Spearman(S), and Intraclass Correlation Coefficients (ICC) between the language model (LM: GPT4-o) and human annotations for the Big 5 personality traits. High ICC values indicate strong agreement between LLM and human evaluations.}
\label{tab:icc}
\end{table*}

\begin{table}[]
\tiny
\centering
\begin{tabular}{p{0.10\linewidth}|p{0.65\linewidth}}
\toprule
\textbf{Role} &
  \textbf{Generated Dialogue} \\
  \midrule
User &
  I struggle to write in my native language. It's like my subconscious is blocking me. \\
Assistant &
  I used to write freely in my journals, but now it feels anxiety-inducing. I miss that liberated feeling. \\
User &
  Exactly! It's like something inside me is making it hard to do deep emotional work. \\
Assistant &
  I think I know what you mean. My writing halted after a surgery, and I haven't found my rhythm again. \\
User &
  Surgery? How did that affect your writing? \\
Assistant &
  Honestly, I thought it was the physical limitation, but even after recovery, the motivation wasn't there. \\
User &
  I get it. Even with my phone, it's hard when I'm in a psychotic episode. Eye strain and stimulation are overwhelming. \\
Assistant &
  I can imagine. It's tough when your mind and body are against you. \\
User &
  Exactly! It feels like an abusive part of me is stopping me from doing the digging work. \\
Assistant &
  I know that feeling. I used to write about everything, no matter how silly or candid. Now, it feels like I've lost that freedom. \\
User &
  Do you think we're just making excuses for ourselves? Maybe we're just scared to face our emotions? \\
Assistant &
  Maybe. But for me, it's more about finding that spark again. I cherish my old journals, but my 30s are void of that creative expression. \\
User &
  I relate. I've been clinically insane for years, and it's hard to even do daily tasks. Writing feels like a luxury I can't afford. \\
Assistant &
  I'm so sorry to hear that. Have you tried any strategies to get back into writing? \\
User &
  Not really. Just trying to find the energy to write. It's hard when your trauma response is so strong. \\
Assistant &
  I get it. For me, it's about rediscovering that joy and urgency to write. Maybe we can support each other in finding our way back? \\
\bottomrule
\end{tabular}
\caption{This Table presents a full dialogue sample annotated with Big 5 personality traits. The scores (ranging from 1 to 5) represent the significance of each trait in the dialogue, with 1 indicating low significance and 5 indicating high significance. The dialogue is evaluated by one language model (GPT4o) and three human judges (J1, J2, J3). The scores for GPT4o are [4, 3, 3, 5, 4], J1 are [4, 2, 2, 5, 5], J2 are [4, 3, 2, 5, 5], and J3 are [4, 3, 2, 5, 4], reflecting an overall agreement in the evaluation of Openness, Conscientiousness, Extraversion, Agreeableness, and Neuroticism (OCEAN).}
\label{tab:quality_eg}
\end{table}


\section{Training Arguments and GPU}
\label{apdx:tr_args}
All the models were trained on a single A100 80 GB. Table \ref{tab:trainer_args} shows the Training Args used to train all the Models. The batch size default was set to 4 but was reduced to 2 when A100 80GB was unavailable (used A100 40 GB). LoRA hyper parameters (r = 64, $alpha$ = 16, dropout = 0.1) were most significant. 

\begin{table}[htbp]
\tiny
\centering
\begin{tabular}{ll}
\toprule
\textbf{Argument}                  & \textbf{Value}                 \\
\midrule
\texttt{lora\_r}                   & 64                             \\
\texttt{lora\_alpha}               & 16                             \\
\texttt{lora\_dropout}             & 0.1                            \\
\texttt{bf16}                      & True                           \\
\texttt{learning\_rate}            & 2.0e-05                        \\
\texttt{gradient\_accumulation\_steps} & 128                        \\
\texttt{gradient\_checkpointing}    & True                           \\
\texttt{logging\_strategy}          & Steps                          \\
\texttt{logging\_steps}             & 1                              \\
\texttt{save\_strategy}             & Steps                          \\
\texttt{save\_steps}                & 100                            \\
\texttt{eval\_steps}                & 100                            \\
\texttt{per\_device\_train\_batch\_size} & 4 \\
\texttt{per\_device\_eval\_batch\_size}  & 4                           \\
\texttt{max\_seq\_length}           & 2048                  \\
\texttt{lr\_scheduler\_type}        & Cosine                         \\
\texttt{early\_stopping\_patience}  & 4                                            \\
\bottomrule
\end{tabular}
\caption{Trainer Arguments}
\label{tab:trainer_args}
\end{table}

\section{Fine-Tuning on Other Datasets}
\label{apdx:ft_other}
We extended the fine-tuning procedure to popular datasets like Persona-Chat (PC), Synthetic Persona-Chat (SPC), and Blended Skill Talk (BST), as described in §\ref{sec:ft_jic}. Let $\mathcal{D}_\text{train}$ represent the training set for each dataset, with sizes $|\mathcal{D}_{\text{PC}}| = 17,878$, $|\mathcal{D}_{\text{SPC}}| = 8,938$, and $|\mathcal{D}_{\text{BST}}| = 4,819$. We used the same parameter-efficient Low-Rank Adaptation (LoRA) method as earlier, updating the projection matrices $W_q$, $W_k$, $W_v$, and $W_o$ while keeping the rest of the model frozen. For the second phase, instead of RAFt, we appended the persona information $\mathcal{P}_i$ from each dataset to the system prompt, resulting in an enriched input $\tilde{x}_i = \{x_i, \mathcal{P}_i\}$. This setup mirrors the RAFt process, where $\mathcal{P}_i$ acts as the additional context, we name it Persona Augmented Fine-tuning (PAFt.). The objective remained to minimize the NLL loss with the persona context guiding the generation to match the personality traits embedded in each dataset.

\section{Detailed Results of Automatic metric-based evaluation}
\label{apdx:automated_detailed}

Table \ref{tab:basic_JIC_detailed} shows detailed results of the automated evaluation on various JIC subsets. Table \ref{tab:basic_other_detailed} shows the results for other datasets on JIC test split.  Table \ref{tab:basic_others} shows the evaluation result of other datasets on their respective test splits.

\begin{table*}[]
\scriptsize
\centering
\begin{tabular}{cccccccccccc}
\toprule
\multirow{2}{*}{\textbf{Dataset}} &
  \multirow{2}{*}{\textbf{Model}} &
  \multirow{2}{*}{\textbf{Train cfg.}} &
  \multirow{2}{*}{\textbf{Test cfg.}} &
  \multirow{2}{*}{\textbf{Train size}} &
  \multicolumn{3}{c}{\textbf{Score}} &
  \multicolumn{3}{c}{\textbf{ROUGE Score}} &
  \multirow{2}{*}{\textbf{AVG}} \\
  \cmidrule(lr){6-8} \cmidrule(lr){9-11}
 &
   &
   &
   &
   &
  \textbf{BLEU} &
  \textbf{METEOR} &
  \textbf{BERT} &
  \textbf{R1} &
  \textbf{R2} &
  \textbf{RL} &
   \\
\midrule
\multirow{4}{*}{JIC-0} &
  \multirow{2}{*}{LLaMA} &
  ZS &
  Reg. &
  - &
  0.2115 &
  0.2176 &
  0.5049 &
  0.2049 &
  0.0646 &
  0.1760 &
  0.2299 \\
 &
   &
  ZS &
  RAG &
  - &
  0.2444 &
  0.2261 &
  0.5210 &
  0.2320 &
  0.0829 &
  0.2033 &
  \textbf{0.2516} \\
  \cmidrule(lr){2-12}
 &
  \multirow{2}{*}{Mistral} &
  ZS &
  Reg. &
  - &
  0.2085 &
  0.2012 &
  0.4957 &
  0.1881 &
  0.0429 &
  0.1559 &
  \textbf{0.2154} \\
 &
   &
  ZS &
  RAG &
  - &
  0.2078 &
  0.1983 &
  0.4912 &
  0.1870 &
  0.0419 &
  0.1545 &
  0.2134 \\
  \cmidrule(lr){1-12}
\multirow{8}{*}{JIC-tiny} &
  \multirow{4}{*}{LLaMA} &
  Ft. &
  Reg. &
  \multirow{4}{*}{6.4k} &
  0.2214 &
  0.2194 &
  0.5092 &
  0.2132 &
  0.0688 &
  0.1850 &
  0.2362 \\
 &
   &
  Ft. &
  RAG &
   &
  0.2462 &
  0.2289 &
  0.5225 &
  0.2342 &
  0.0849 &
  0.2054 &
  0.2537 \\
 &
   &
  RAFt. &
  Reg. &
   &
  0.2296 &
  0.2206 &
  0.5146 &
  0.2234 &
  0.0758 &
  0.1953 &
  0.2432 \\
 &
   &
  RAFt. &
  RAG &
   &
  0.2477 &
  0.2305 &
  0.5244 &
  0.2399 &
  0.0889 &
  0.2115 &
  \textbf{0.2571} \\
  \cmidrule(lr){2-12}
 &
  \multirow{4}{*}{Mistral} &
  Ft. &
  Reg. &
  \multirow{4}{*}{6.4k} &
  0.2078 &
  0.1858 &
  0.4864 &
  0.1846 &
  0.0398 &
  0.1544 &
  0.2098 \\
 &
   &
  Ft. &
  RAG &
   &
  0.2096 &
  0.1927 &
  0.4911 &
  0.1872 &
  0.0425 &
  0.1549 &
  \textbf{0.2130} \\
 &
   &
  RAFt. &
  Reg. &
   &
  0.2031 &
  0.1762 &
  0.4801 &
  0.1770 &
  0.0365 &
  0.1469 &
  0.2033 \\
 &
   &
  RAFt. &
  RAG &
   &
  0.2055 &
  0.1848 &
  0.4846 &
  0.1812 &
  0.0378 &
  0.1490 &
  0.2071 \\
  \cmidrule(lr){1-12}
\multirow{8}{*}{JIC-small} &
  \multirow{4}{*}{LLaMA} &
  Ft. &
  Reg. &
  \multirow{4}{*}{19k} &
  0.2313 &
  0.2235 &
  0.5156 &
  0.2234 &
  0.0750 &
  0.1951 &
  0.2440 \\
 &
   &
  Ft. &
  RAG &
   &
  0.2444 &
  0.2261 &
  0.5210 &
  0.2320 &
  0.0830 &
  0.2033 &
  0.2516 \\
 &
   &
  RAFt. &
  Reg. &
   &
  0.2471 &
  0.2351 &
  0.5264 &
  0.2425 &
  0.0855 &
  0.2138 &
  \textbf{0.2584} \\
 &
   &
  RAFt. &
  RAG &
   &
  0.2488 &
  0.2372 &
  0.5190 &
  0.2407 &
  0.0920 &
  0.2120 &
  0.2583 \\
  \cmidrule(lr){2-12}
 &
  \multirow{4}{*}{Mistral} &
  Ft. &
  Reg. &
  \multirow{4}{*}{19k} &
  0.2079 &
  0.1872 &
  0.4881 &
  0.1879 &
  0.0417 &
  0.1582 &
  0.2118 \\
 &
   &
  Ft. &
  RAG &
   &
  0.2078 &
  0.1983 &
  0.4912 &
  0.1870 &
  0.0419 &
  0.1546 &
  \textbf{0.2135} \\
 &
   &
  RAFt. &
  Reg. &
   &
  0.2070 &
  0.1814 &
  0.4829 &
  0.1823 &
  0.0395 &
  0.1523 &
  0.2075 \\
 &
   &
  RAFt. &
  RAG &
   &
  0.2089 &
  0.1887 &
  0.4863 &
  0.1852 &
  0.0415 &
  0.1530 &
  0.2106 \\
  \cmidrule(lr){1-12}
\multirow{8}{*}{JIC-medium} &
  \multirow{4}{*}{LLaMA} &
  Ft. &
  Reg. &
  \multirow{4}{*}{29k} &
  0.2391 &
  0.2294 &
  0.5215 &
  0.2320 &
  0.0803 &
  0.2036 &
  0.2510 \\
 &
   &
  Ft. &
  RAG &
   &
  0.2444 &
  0.2261 &
  0.5210 &
  0.2320 &
  0.0830 &
  0.2033 &
  0.2517 \\
 &
   &
  RAFt. &
  Reg. &
   &
  0.2520 &
  0.2401 &
  0.5283 &
  0.2458 &
  0.0871 &
  0.2166 &
  \textbf{0.2616} \\
 &
   &
  RAFt. &
  RAG &
   &
  0.2511 &
  0.2412 &
  0.5214 &
  0.2433 &
  0.0927 &
  0.2142 &
  0.2606 \\
  \cmidrule(lr){2-12}
 &
  \multirow{4}{*}{Mistral} &
  Ft. &
  Reg. &
  \multirow{4}{*}{29k} &
  0.2189 &
  0.1961 &
  0.4973 &
  0.1974 &
  0.0477 &
  0.1672 &
  \textbf{0.2208} \\
 &
   &
  Ft. &
  RAG &
   &
  0.2078 &
  0.1983 &
  0.4912 &
  0.1870 &
  0.0419 &
  0.1545 &
  0.2134 \\
 &
   &
  RAFt. &
  Reg. &
   &
  0.2130 &
  0.1858 &
  0.4870 &
  0.1881 &
  0.0434 &
  0.1579 &
  0.2125 \\
 &
   &
  RAFt. &
  RAG &
   &
  0.2183 &
  0.1943 &
  0.4922 &
  0.1937 &
  0.0460 &
  0.1609 &
  0.2176 \\
  \cmidrule(lr){1-12}
\multirow{4}{*}{JIC-large} &
  \multirow{2}{*}{LLaMA} &
  Ft. &
  Reg. &
  \multirow{2}{*}{99k} &
  0.2619 &
  0.2559 &
  0.5356 &
  0.2550 &
  0.0980 &
  0.2248 &
  \textbf{0.2719} \\
 &
   &
  Ft. &
  RAG &
   &
  0.2503 &
  0.2495 &
  0.5229 &
  0.2468 &
  0.1013 &
  0.2184 &
  0.2649 \\
  \cmidrule(lr){2-12}
 &
  \multirow{2}{*}{Mistral} &
  Ft. &
  Reg. &
  \multirow{2}{*}{99k} &
  0.2316 &
  0.2112 &
  0.5082 &
  0.2116 &
  0.0561 &
  0.1813 &
  0.2333 \\
 &
   &
  Ft. &
  RAG &
   &
  0.2381 &
  0.2212 &
  0.5132 &
  0.2199 &
  0.0613 &
  0.1874 &
  \textbf{0.2402} \\
  \cmidrule(lr){1-12}
\multirow{8}{*}{JIC-$\alpha_{1}\beta_{0}$} &
  \multirow{4}{*}{LLaMA} &
  Ft. &
  Reg. &
  \multirow{4}{*}{$\sim$135k} &
  0.2743 &
  0.2620 &
  0.5436 &
  0.2657 &
  0.1053 &
  0.2360 &
  0.2812 \\
 &
   &
  Ft. &
  RAG &
   &
  0.2444 &
  0.2261 &
  0.5210 &
  0.2319 &
  0.0830 &
  0.2034 &
  0.2516 \\
 &
   &
  RAFt. &
  Reg. &
   &
  0.2722 &
  0.2595 &
  0.5418 &
  0.2625 &
  0.0992 &
  0.2319 &
  0.2778 \\
 &
   &
  RAFt. &
  RAG &
   &
  0.2778 &
  0.2680 &
  0.5436 &
  0.2702 &
  0.1073 &
  0.2390 &
  \textbf{0.2843} \\
  \cmidrule(lr){2-12}
 &
  \multirow{4}{*}{Mistral} &
  Ft. &
  Reg. &
  \multirow{4}{*}{$\sim$135k} &
  0.2371 &
  0.2182 &
  0.5125 &
  0.2186 &
  0.0600 &
  0.1872 &
  0.2389 \\
 &
   &
  Ft. &
  RAG &
   &
  0.2430 &
  0.2270 &
  0.5181 &
  0.2261 &
  0.0642 &
  0.1931 &
  \textbf{0.2453} \\
 &
   &
  RAFt. &
  Reg. &
   &						
  0.2149 &
  0.1985 &
  0.4934 &
  0.1972 &
  0.0486 &
  0.1669 &
  0.2199 \\
 &
   &
  RAFt. &
  RAG &
   &
  0.2364 &
  0.2108 &
  0.5088 &
  0.2136 &
  0.0567 &
  0.1822 &
  0.2348 \\
  \cmidrule(lr){1-12}
\multirow{4}{*}{JIC-all} &
  \multirow{2}{*}{LLaMA} &
  Ft. &
  Reg. &
  \multirow{2}{*}{$\sim$400k} &
  0.2967 &
  0.2826 &
  0.5587 &
  0.2878 &
  0.1210 &
  0.2576 &
  0.3007 \\
 &
   &
  Ft. &
  RAG &
   &
  0.3062 &
  0.2945 &
  0.5651 &
  0.2989 &
  0.1308 &
  0.2676 &
  \underline{\textbf{0.3105}} \\
  \cmidrule(lr){2-12}
 &
  \multirow{2}{*}{Mistral} &
  Ft. &
  Reg. &
  \multirow{2}{*}{$\sim$400k} &
  0.2589 &
  0.2382 &
  0.5273 &
  0.2392 &
  0.0763 &
  0.2077 &
  0.2579 \\
 &
   &
  Ft. &
  RAG &
   &
  0.2665 &
  0.2451 &
  0.5334 &
  0.2468 &
  0.0813 &
  0.2146 &
  \textbf{0.2646} \\
\bottomrule
\end{tabular}
\caption{Evaluation of LLaMA 3 8B Instruct and Mistral 7B v0.3 Instruct models on the JIC dataset, using various configurations and dataset subsets. The training configurations include Zero-shot (ZS), Fine-tuning on dialogues (Ft.), and Retrieval-augmented Fine-tuning on dialogues (RAFt.). Inference was performed with regular (Reg.) or Retrieval Augmented Generation (RAG) settings. The validation split and test split for the evaluation were kept constant at 1k and 2k samples, respectively. The highest average score across models for each dataset subset is highlighted. The best score in the Table is underlined.}
\label{tab:basic_JIC_detailed}
\end{table*}

\begin{table*}[]
\scriptsize
\centering
\begin{tabular}{ccccccccccc}
\toprule
\multirow{2}{*}{\textbf{Dataset}} &
  \multirow{2}{*}{\textbf{Model}} &
  \multirow{2}{*}{\textbf{Tr. cfg.}} &
  \multirow{2}{*}{\textbf{Ts. cfg.}} &
  \multicolumn{3}{c}{\textbf{Score}} &
  \multicolumn{3}{c}{\textbf{ROUGE Score}} &
  \multirow{2}{*}{\textbf{AVG}} \\
  \cmidrule(lr){5-7} \cmidrule(lr){8-10}
  
                     &                          &       &      & \textbf{BLEU} & \textbf{METEOR} & \textbf{BERT} & \textbf{R1} & \textbf{R2} & \textbf{RL} &        \\
                     \midrule
\multirow{4}{*}{PC}  & \multirow{2}{*}{LLaMA}   & PAFt. & Reg. & 0.2228        & 0.2191          & 0.5103        & 0.2151      & 0.0699      & 0.1868      & 0.2373 \\
                     &                          & PAFt. & RAG  & 0.2454        & 0.2282          & 0.5227        & 0.2353      & 0.0849      & 0.2065      & 0.2538 \\
                      \cmidrule(lr){2-11}
                     & \multirow{2}{*}{Mistral} & PAFt. & Reg. & 0.2073        & 0.1842          & 0.4851        & 0.1821      & 0.0390      & 0.1522      & 0.2083 \\
                     &                          & PAFt. & RAG  & 0.2087        & 0.1929          & 0.4892        & 0.1862      & 0.0416      & 0.1548      & 0.2122 \\
                     \midrule
\multirow{4}{*}{SPC} & \multirow{2}{*}{LLaMA}   & PAFt. & Reg. & 0.2237        & 0.2209          & 0.5108        & 0.2160      & 0.0706      & 0.1874      & 0.2382 \\
                     &                          & PAFt. & RAG  & 0.2464        & 0.2291          & 0.5233        & 0.2358      & 0.0847      & 0.2071      & 0.2544 \\
                     \cmidrule(lr){2-11}
                     & \multirow{2}{*}{Mistral} & PAFt. & Reg. & 0.2086        & 0.1847          & 0.4864        & 0.1836      & 0.0402      & 0.1540      & 0.2096 \\
                     &                          & PAFt. & RAG  & 0.2102        & 0.1929          & 0.4905        & 0.1874      & 0.0423      & 0.1560      & 0.2132 \\
                     \midrule
\multirow{4}{*}{BST} & \multirow{2}{*}{LLaMA}   & PAFt. & Reg. & 0.2203        & 0.2188          & 0.5083        & 0.2120      & 0.0680      & 0.1837      & 0.2352 \\
                     &                          & PAFt. & RAG  & 0.2443        & 0.2262          & 0.5212        & 0.2322      & 0.0831      & 0.2036      & 0.2518 \\
                     \cmidrule(lr){2-11}
                     & \multirow{2}{*}{Mistral} & PAFt. & Reg. & 0.2084        & 0.1881          & 0.4869        & 0.1838      & 0.0402      & 0.1533      & 0.2101 \\
                     &                          & PAFt. & RAG  & 0.2078        & 0.1977          & 0.4911        & 0.1872      & 0.0419      & 0.1548      & 0.2134 \\
                     \bottomrule
\end{tabular}
\caption{Evaluation of LLaMA 3 8B Instruct and Mistral 7B v0.3 Instruct models on other Datasets, using various configurations. The training configurations include Persona Augmented Fine-tuning (PAFt.) on the full dataset. Inference was performed with regular (Reg.) or Retrieval Augmented Generation (RAG) settings. Test splits for the evaluation were kept constant at 2k samples from JIC.}
\label{tab:basic_other_detailed}
\end{table*}

\begin{table*}[]
\scriptsize
\centering
\begin{tabular}{cccccccccc}
\toprule
\multirow{2}{*}{\textbf{Dataset}} &
  \multirow{2}{*}{\textbf{Model}} &
  \multirow{2}{*}{\textbf{Train cfg.}} &
  \multicolumn{3}{c}{\textbf{Score}} &
  \multicolumn{3}{c}{\textbf{ROUGE}} &
  \multirow{2}{*}{\textbf{AVG}} \\
  \cmidrule(lr){4-6} \cmidrule(lr){7-9}
                     &                          &      & \textbf{BLEU} & \textbf{METEOR} & \textbf{BERT} & \textbf{R1} & \textbf{R2} & \textbf{RL} &                 \\
                     \midrule
\multirow{6}{*}{PC}  & \multirow{3}{*}{LLaMA}   & ZS   & 0.1183        & 0.1419          & 0.4272        & 0.1268      & 0.0207      & 0.1122      & 0.1579          \\
                     &                          & Ft.  & 0.1275        & 0.1383          & 0.4318        & 0.1345      & 0.0235      & 0.1202      & 0.1626          \\
                     &                          & PAFt. & 0.1592        & 0.1608          & 0.4550        & 0.1675      & 0.0378      & 0.1497      & \underline{\textbf{0.1883}} \\
                     \cmidrule(lr){2-10}
                     & \multirow{3}{*}{Mistral} & ZS   & 0.1340        & 0.1564          & 0.4314        & 0.1364      & 0.0206      & 0.1199      & 0.1665          \\
                     &                          & Ft.  & 0.1367        & 0.1276          & 0.4247        & 0.1310      & 0.0183      & 0.1171      & 0.1592          \\
                     &                          & PAFt. & 0.1551        & 0.1537          & 0.4419        & 0.1489      & 0.0270      & 0.1320      & \textbf{0.1764} \\
                     \midrule
\multirow{6}{*}{SPC} & \multirow{3}{*}{LLaMA}   & ZS   & 0.1962        & 0.2806          & 0.5369        & 0.2542      & 0.1165      & 0.2386      & 0.2705          \\
                     &                          & Ft.  & 0.2093        & 0.2870          & 0.5458        & 0.2675      & 0.1296      & 0.2519      & 0.2819          \\
                     &                          & PAFt. & 0.2415        & 0.3080          & 0.5759        & 0.3079      & 0.1629      & 0.2936      & \underline{\textbf{0.3150}} \\
                     \cmidrule(lr){2-10}
                     & \multirow{3}{*}{Mistral} & ZS   & 0.2041        & 0.3199          & 0.5462        & 0.2586      & 0.1154      & 0.2378      & 0.2803          \\
                     &                          & Ft.  & 0.2452        & 0.3208          & 0.5704        & 0.3006      & 0.1520      & 0.2837      & \textbf{0.3121} \\
                     &                          & PAFt. & 0.2358        & 0.3184          & 0.5585        & 0.2861      & 0.1391      & 0.2664      & 0.3007          \\
                     \midrule
\multirow{6}{*}{BST} & \multirow{3}{*}{LLaMA}   & ZS   & 0.1370        & 0.1437          & 0.4379        & 0.1378      & 0.0220      & 0.1145      & 0.1655          \\
                     &                          & Ft.  & 0.1372        & 0.1431          & 0.4379        & 0.1374      & 0.0220      & 0.1143      & 0.1653          \\
                     &                          & PAFt. & 0.1475        & 0.1395          & 0.4491        & 0.1541      & 0.0294      & 0.1319      & \underline{\textbf{0.1752}} \\
                     \cmidrule(lr){2-10}
                     & \multirow{3}{*}{Mistral} & ZS   & 0.1434        & 0.1357          & 0.4320        & 0.1355      & 0.0185      & 0.1130      & 0.1630          \\
                     &                          & Ft.  & 0.1446        & 0.1399          & 0.4334        & 0.1374      & 0.0190      & 0.1132      & 0.1646          \\
                     &                          & PAFt. & 0.1486        & 0.1459          & 0.4363        & 0.1396      & 0.0190      & 0.1155      & \textbf{0.1675} \\
\bottomrule
\end{tabular}
\caption{Evaluation of LLaMA 3 8B Instruct and Mistral 7B v0.3 Instruct models on three datasets: Persona Chat (PC), Synthetic Persona Chat (SPC), and Blended Skill Talk (BST), using various training configurations. The training configurations include zero-shot (ZS), fine-tuning on dialogues (Ft.), and persona-augmented finetuning (PAFt.). The entire dataset was used for each training configuration. Testing was performed on their respective test splits. The highest average score across models for each dataset subset is highlighted, and the best overall score for each dataset in the Table is underlined.}
\label{tab:basic_others}
\end{table*}

\section{Detailed Results of LM-Eval Harness}
\label{apdx:lm-eval-detailed}

Table \ref{tab:lm_eval_main_detail} shows the scores for Big 5 traits (O,C,E,A,N) in JIC subsets. Table \ref{tab:lm_eval_others} shows the Big 5 traits in other datasets.

\begin{table*}[]
\scriptsize
\centering
\begin{tabular}{cccccccccc}
\toprule
\multirow{2}{*}{\textbf{Dataset}} &
  \multirow{2}{*}{\textbf{Model}} &
  \multirow{2}{*}{\textbf{Train. cfg.}} &
  \multirow{2}{*}{\textbf{Train size}} &
  \multicolumn{5}{c}{\textbf{Personality Traits Score}} &
  \multirow{2}{*}{\textbf{AVG}} \\
  \cmidrule(lr){5-9}
 &
   &
   &
   &
  \textbf{O} &
  \textbf{C} &
  \textbf{E} &
  \textbf{A} &
  \textbf{N} &
   \\
   \midrule
\multirow{2}{*}{-} &
  LLaMA &
  ZS &
  - &
  0.8760 &
  0.7620 &
  0.7170 &
  0.9500 &
  0.5220 &
  0.7654 \\
  \cmidrule(lr){2-10}
 &
  Mistral &
  ZS &
  - &
  0.8360 &
  0.6390 &
  0.5140 &
  0.8160 &
  0.5350 &
  0.6680 \\
  \midrule
\multirow{4}{*}{JIC-tiny} &
  \multirow{2}{*}{LLaMA} &
  Ft. &
  6.4k &
  0.8770 &
  0.7660 &
  0.7180 &
  0.9500 &
  0.5280 &
  0.7678 \\
 &
   &
  RAFt. &
  6.4k &
  0.8780 &
  0.7750 &
  0.7210 &
  0.9510 &
  0.5320 &
  0.7714 \\
  \cmidrule(lr){2-10}
 &
  \multirow{2}{*}{Mistral} &
  Ft. &
  6.4k &
  0.8380 &
  0.6460 &
  0.5140 &
  0.8160 &
  0.5350 &
  0.6698 \\
 &
   &
  RAFt. &
  6.4k &
  0.8500 &
  0.6540 &
  0.5130 &
  0.8260 &
  0.5370 &
  0.6760 \\
  \midrule
\multirow{4}{*}{JIC-small} &
  \multirow{2}{*}{LLaMA} &
  Ft. &
  19k &
  0.8750 &
  0.7750 &
  0.7180 &
  0.9510 &
  0.5290 &
  0.7696 \\
 &
   &
  RAFt. &
  19k &
  0.8890 &
  0.8000 &
  0.7390 &
  0.9550 &
  0.5330 &
  0.7832 \\
  \cmidrule(lr){2-10}
 &
  \multirow{2}{*}{Mistral} &
  Ft. &
  19k &
  0.8540 &
  0.6600 &
  0.5150 &
  0.8280 &
  0.5370 &
  0.6788 \\
 &
   &
  RAFt. &
  19k &
  0.8830 &
  0.6970 &
  0.5340 &
  0.8640 &
  0.5410 &
  0.7038 \\
  \midrule
\multirow{4}{*}{JIC-medium} &
  \multirow{2}{*}{LLaMA} &
  Ft. &
  29k &
  0.8770 &
  0.7800 &
  0.7160 &
  0.9520 &
  0.5300 &
  0.7710 \\
 &
   &
  RAFt. &
  29k &
  0.8980 &
  \textbf{0.8040} &
  {\ul \textbf{0.7400}} &
  0.9600 &
  0.5350 &
  0.7874 \\
  \cmidrule(lr){2-10}
 &
  \multirow{2}{*}{Mistral} &
  Ft. &
  29k &
  0.8670 &
  0.6820 &
  0.5220 &
  0.8600 &
  0.5380 &
  0.6938 \\
 &
   &
  RAFt. &
  29k &
  0.9070 &
  0.7330 &
  0.5790 &
  0.8940 &
  0.5450 &
  0.7316 \\
  \midrule
\multirow{8}{*}{JIC ($\alpha$,$\beta$)} &
  \multirow{4}{*}{LLaMA} &
  Ft. $\alpha_{0}\beta_{0}$ &
  $\sim$100k &
  0.8810 &
  0.7980 &
  0.7140 &
  0.9580 &
  0.5380 &
  0.7778 \\
 &
   &
  Ft. $\alpha_{1}\beta_{0}$ &
  $\sim$135k &
  0.8860 &
  0.7930 &
  0.7080 &
  {\ul \textbf{0.9600}} &
  0.5570 &
  0.7808 \\
 &
   &
  \textbf{RAFt. $\alpha_{1}\beta_{0}$*} &
  $\sim$135k &
  \textbf{0.9150} &
  0.7840 &
  0.6680 &
  0.9440 &
  {\ul \textbf{0.7040}} &
  {\ul \textbf{0.8030}} \\
 &
   &
  Ft. $\alpha_{1}\beta_{0.5}$ &
  $\sim$220k &
  0.8830 &
  0.7990 &
  0.7080 &
  0.9580 &
  0.5380 &
  0.7772 \\
  \cmidrule(lr){2-10}
 &
  \multirow{4}{*}{Mistral} &
  Ft. $\alpha_{0}\beta_{0}$ &
  $\sim$100k &
  0.9090 &
  0.7430 &
  0.5690 &
  0.9030 &
  0.5490 &
  0.7346 \\
 &
   &
  Ft. $\alpha_{1}\beta_{0}$ &
  $\sim$135k &
  0.9120 &
  0.7480 &
  0.5800 &
  0.9090 &
  0.5490 &
  0.7396 \\
 &
   &
  \textbf{RAFt. $\alpha_{1}\beta_{0}$*} &
  $\sim$135k &
  {\ul \textbf{0.9340}} &
  {\ul \textbf{0.8260}} &
  \textbf{0.6250} &
  \textbf{0.9530} &
  \textbf{0.5700} &
  \textbf{0.7816} \\
 &
   &
  Ft. $\alpha_{1}\beta_{0.5}$ &
  $\sim$220k &
  0.9050 &
  0.7530 &
  0.5760 &
  0.9140 &
  0.5590 &
  0.7414 \\
  \midrule
\multirow{2}{*}{JIAC-all} &
  LLaMA &
  Ft. &
  $\sim$400k &
  0.8340 &
  0.6990 &
  0.6000 &
  0.9260 &
  0.6450 &
  0.7408 \\
  \cmidrule(lr){2-10}
 &
  Mistral &
  Ft. &
  $\sim$400k &
  0.8840 &
  0.7110 &
  0.5310 &
  0.9110 &
  0.5650 &
  0.7204 \\
\bottomrule
\end{tabular}
\caption{Evaluation of the Personality Trait Scores using LM Eval Harness across various subsets of the Journal Intensive Conversation (JIC) dataset using LLaMA 3 8B Instruct and Mistral 7B v0.3 Instruct models. The training configurations include Zero-shot (ZS), Fine-tuning on dialogues (Ft.), and Retrieval-augmented Fine-tuning on dialogues (RAFt.). For each model and subset, the scores for the Big Five personality traits—Openness (O), Conscientiousness (C), Extraversion (E), Agreeableness (A), and Neuroticism (N)—were computed. The highest scores within each model are emphasized, and the best overall score per trait is underlined. * represents our best configuration.}
\label{tab:lm_eval_main_detail}
\end{table*}

\begin{table*}[]
\scriptsize
\centering
\begin{tabular}{ccccccccc}
\toprule
\multirow{2}{*}{\textbf{Dataset}} &
  \multirow{2}{*}{\textbf{Model}} &
  \multirow{2}{*}{\textbf{Train. cfg.}} &
  \multicolumn{5}{c}{\textbf{Personality Traits Score}} &
  \multirow{2}{*}{\textbf{AVG}} \\
  \cmidrule(lr){4-8}
                     &                          &      & \textbf{O} & \textbf{C} & \textbf{E} & \textbf{A} & \textbf{N} &        \\
                     \midrule
\multirow{4}{*}{PC}  & \multirow{2}{*}{LLaMA}   & Ft.  & 0.8740     & 0.7660     & 0.7180     & 0.9510     & 0.5240     & 0.7666 \\
                     &                          & PAFt. & 0.8750     & 0.7650     & 0.7140     & 0.9490     & 0.5250     & 0.7656 \\
                     \cmidrule(lr){2-9}
                     & \multirow{2}{*}{Mistral} & Ft.  & 0.8380     & 0.6380     & 0.5140     & 0.8080     & 0.5370     & 0.6670 \\
                     &                          & PAFt. & 0.8310     & 0.6350     & 0.5110     & 0.8050     & 0.5370     & 0.6638 \\
                     \midrule
\multirow{4}{*}{SPC} & \multirow{2}{*}{LLaMA}   & Ft.  & 0.8760     & 0.7620     & 0.7150     & 0.9480     & 0.5260     & 0.7654 \\
                     &                          & PAFt. & 0.8750     & 0.7680     & 0.7190     & 0.9500     & 0.5240     & 0.7672 \\
                     \cmidrule(lr){2-9}
                     & \multirow{2}{*}{Mistral} & Ft.  & 0.8280     & 0.6340     & 0.5120     & 0.8050     & 0.5360     & 0.6630 \\
                     &                          & PAFt. & 0.8320     & 0.6330     & 0.5120     & 0.8100     & 0.5360     & 0.6646 \\
                     \midrule
\multirow{4}{*}{BST} & \multirow{2}{*}{LLaMA}   & Ft.  & 0.8760     & 0.7600     & 0.7170     & 0.9500     & 0.5210     & 0.7648 \\
                     &                          & PAFt. & 0.8760     & 0.7630     & 0.7160     & 0.9510     & 0.5240     & 0.7660 \\
                     \cmidrule(lr){2-9}
                     & \multirow{2}{*}{Mistral} & Ft.  & 0.8360     & 0.6400     & 0.5130     & 0.8180     & 0.5340     & 0.6682 \\
                     &                          & PAFt. & 0.8350     & 0.6400     & 0.5140     & 0.8130     & 0.5360     & 0.6676 \\
                     \bottomrule
\end{tabular}
\caption{Evaluation of Personality Trait Scores on external datasets including Persona Chat (PC), Synthetic Persona Chat (SPC), and Blended Skill Talk (BST), using LLaMA 3 8B Instruct and Mistral 7B v0.3 Instruct models. The training configurations include Fine-tuning on dialogues (Ft.) and Persona Augmented Fine-tuning (PAFt.). The personality traits evaluated include Openness (O), Conscientiousness (C), Extraversion (E), Agreeableness (A), and Neuroticism (N). Results show comparable performance to the baseline models across the datasets, with no significant improvements in personality trait representation.}
\label{tab:lm_eval_others}
\end{table*}

\section{Qualitative Analysis Examples}
\label{apdx:q_eg}

Table \ref{tab:q001} compares real and model-generated dialogues capturing personality traits.
\begin{table}[]
\tiny
\centering
\begin{tabular}{>{\centering\arraybackslash}p{0.15\linewidth}|p{0.10\linewidth}p{0.4\linewidth}p{0.15\linewidth}}
\toprule
\textbf{Settings} &
  \multicolumn{1}{c}{\textbf{Role}} &
  \multicolumn{1}{c}{\textbf{Prompt \& Response}} &
  \textbf{Captured Traits} \\
  \midrule
\multirow{7}{*}{Prompt} &
  User &
  Hey, I \textcolor{RoyalBlue}{created an app} for simple journaling. \textcolor{RoyalBlue}{No frills, just a stream of logs}. &
  \textcolor{RoyalBlue}{Openness} \\
 &
  Assistant &
  Hmm, interesting. I've been dealing with \textcolor{BrickRed}{family drama}, not much time for journaling. &
  \textcolor{BrickRed}{Neuroticism} \\
 &
  User &
  \textcolor{ForestGreen}{Sorry to hear that. Family can be tough}. What's going on with your family? &
  \textcolor{ForestGreen}{Agreeableness} \\
 &
  Assistant &
  They're \textcolor{BrickRed}{judging my clothes and saying I'm disrespectful. It hurts}. &
  \textcolor{BrickRed}{Neuroticism} \\
 &
  User &
  \textcolor{ForestGreen}{That must be tough}. I created ZenJournal to escape distractions, \textcolor{RoyalBlue}{maybe it can help you too}? &
  \textcolor{ForestGreen}{Agreeableness}, \textcolor{RoyalBlue}{Openness} \\
 &
  Assistant &
  Maybe. But honestly, \textcolor{TealBlue}{I just wish my family would talk to me directly} instead of gossiping. &
  \textcolor{TealBlue}{Conscientiousness} \\
 &
  User &
  \textcolor{TealBlue}{Direct communication is key}. I wanted my app to be \textcolor{RoyalBlue}{stress-free, no ceremony, just writing}. &
  \textcolor{TealBlue}{Conscientiousness}, \textcolor{RoyalBlue}{Openness} \\
  \midrule
Golden &
  Assistant &
  Exactly! I feel like my \textcolor{BrickRed}{family is making a big deal out of nothing, and it's stressful}. &
  \textcolor{BrickRed}{Neuroticism} \\
  \midrule
\textbf{LLaMA (our best)} &
  Assistant &
  Yeah, I wish my \textcolor{BrickRed}{family would listen to me without interrupting}. &
  \textcolor{BrickRed}{Neuroticism} \\
  \midrule
  LLaMA PAFt. SPC &
  Assistant &
  \textcolor{ForestGreen}{That's really refreshing. Sometimes simplicity is the best way to get things off your chest.}. &
  \textcolor{ForestGreen}{Agreeableness} \\
  \midrule
LLaMA (base) &
  Assistant &
  \textcolor{ForestGreen}{That's really refreshing. Sometimes simplicity is the best way to go}. &
  \textcolor{ForestGreen}{Agreeableness} \\
  \midrule
\textbf{Mistral (our best)} & Assistant & That's smart. I've been so \textcolor{BrickRed}{busy with family conflict}, I haven't had time to reflect.                 & \textcolor{BrickRed}{Neuroticism}             \\
\midrule
Mistral (base) \& PAFt. SPC              & Assistant & That's a \textcolor{ForestGreen}{good approach}. \textcolor{RoyalBlue}{I think I'll try ZenJournal}. Maybe I can explain how I feel without anxiety. & \textcolor{ForestGreen}{Agreeableness}, \textcolor{RoyalBlue}{Openness} \\
\bottomrule
\end{tabular}
\caption{Comparison of real and model-generated dialogues capturing personality traits. The Table demonstrates how our best-performing models (LLaMA and Mistral) align with the traits reflected in the original dialogue. The captured traits, such as Openness, Neuroticism, Agreeableness, and Conscientiousness, are listed for each utterance to highlight the model's ability to replicate real conversational nuances.}
\label{tab:q001}
\end{table}

\section{Test Set Analysis}
\label{apdx:test_analysis}

The test set analysis was conducted to evaluate dialogue-specific Big 5 personality traits, and the following percentages were observed for each trait, given in Table \ref{tab:big5_percentages}.

\begin{table}[htbp]
\tiny
\centering
\begin{tabular}{ll}
\toprule
\textbf{Trait}                  & \textbf{Percentage (\%)}        \\
\midrule
\texttt{Agreeableness}          & 98.76                          \\
\texttt{Openness}               & 100.00                         \\
\texttt{Conscientiousness}      & 54.77                          \\
\texttt{Extraversion}           & 90.71                          \\
\texttt{Neuroticism}            & 92.56                          \\
\bottomrule
\end{tabular}
\caption{Observed Big 5 Personality Trait Percentages in the Test Set}
\label{tab:big5_percentages}
\end{table}

These results indicate that the test set predominantly exhibits high levels of all traits except \emph{Conscientiousness}, which shows a more balanced distribution. This suggests that the dataset is skewed towards dialogues characterized by predominant traits, potentially impacting the generalizability of predictions for conscientiousness.

\section{Toxicity Analysis Statistics}
\label{apdx:toxic}

\begin{figure}[htbp]
    \centering
    \resizebox{0.5\textwidth}{!}{\includesvg{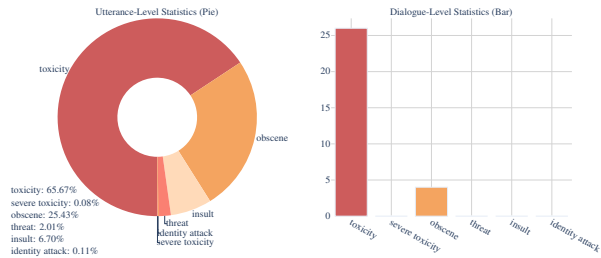}}
    \caption{Left: Utterance-Level Pie Chart(around 11k utt.) and Right: Dialogue-Level Bar Chart Toxicity Subclass Counts}
    \label{fig:toxic}
\end{figure}

Fig. \ref{fig:toxic} shows the detailed statistics for the toxicity subclasses.

Table \ref{tab:toxicity_analysis} shows the exact number of utterances and dialogues classified as toxic.

\begin{table}[htbp]
\scriptsize
\centering
\begin{tabular}{lcc}
\toprule
\textbf{Subclass}        & \textbf{Utterance-level} & \textbf{Dialogue-level} \\
\midrule
Toxicity                 & 10,871                  & 26                     \\
Severe Toxicity          & 13                      & 0                      \\
Obscene                  & 4,210                   & 4                      \\
Threat                   & 333                     & 0                      \\
Insult                   & 1,110                   & 0                      \\
Identity Attack          & 18                      & 0                      \\
\bottomrule
\end{tabular}
\caption{Toxicity analysis showing counts of utterances and dialogues classified under each subclass.}
\label{tab:toxicity_analysis}
\end{table}

\section{LM-Eval Harness Common Tasks}
\label{apdx:lm_eval_other_bench}

Table \ref{tab:lm_eval_other_bench} shows how the model fine-tuned in various JIC splits performed in average on various popular benchmarks to assess general reasoning, knowledge reasoning, and reading comprehension.
\begin{table*}[]
\tiny
\centering
\begin{tabular}{cccccccccccccc}
\toprule
\multirow{2}{*}{\textbf{Tr. cfg.}} &
  \multicolumn{5}{c}{\textbf{General}} &
  \multicolumn{5}{c}{\textbf{Knowledge Reasoning}} &
  \multicolumn{2}{c}{\textbf{Reading Comp.}} &
  \multicolumn{1}{l}{\multirow{2}{*}{\textbf{AVG}}} \\
  \cmidrule(lr){2-6} \cmidrule(lr){7-11} \cmidrule(lr){12-13}
   &
  \textbf{MMLU(0)} &
  \textbf{AGIev.} &
  \textbf{CSQA} &
  \textbf{wg} &
  \textbf{ARC} &
  \textbf{GPQA(0)} &
  \textbf{TfQA1} &
  \textbf{TfQA2} &
  \textbf{TfQA(g)} &
  \textbf{TrQA} &
  \textbf{BoolQ} &
  \textbf{hs} &
  \multicolumn{1}{l}{} \\
  \midrule
ZS (L)                   & 0.6389 & 0.3577 & 0.7592 & 0.7206 & 0.5265 & 0.2969 & 0.3611 & 0.5164 & 0.4676 & 0.5111 & 0.8306 & 0.5770 & 0.5469 \\
                           \cmidrule(lr){1-14}
ZS (M)                   & 0.5978 & 0.3245 & 0.6937 & 0.7419 & 0.5725 & 0.3170 & 0.4235 & 0.5966 & 0.5692 & 0.5673 & 0.8584 & 0.6477 & 0.5758 \\
                           \midrule
$\alpha_{0}\beta_{0}$ (L)             & 0.6324 & 0.3528 & 0.7592 & 0.7167 & 0.5222 & 0.3036 & 0.3390 & 0.5056 & 0.5214 & 0.5186 & 0.8327 & 0.5723 & 0.5480 \\
$\alpha_{1}\beta_{0}$ (L)             & 0.6319 & 0.3518 & 0.7543 & 0.7174 & 0.5256 & 0.3192 & 0.3427 & 0.5062 & 0.5214 & 0.5185 & 0.8333 & 0.5711 & 0.5495 \\
\textbf{R. $\alpha_{1}\beta_{0}$* (L)} & 0.6283 & 0.3463 & 0.7535 & 0.7301 & 0.5282 & 0.2991 & 0.3341 & 0.5005 & 0.5838 & 0.5361 & 0.8321 & 0.5722 & 0.5537 \\
$\alpha_{1}\beta_{0.5}$ (L)           & 0.6318 & 0.3533 & 0.7568 & 0.7214 & 0.5230 & 0.3080 & 0.3403 & 0.5049 & 0.5251 & 0.5208 & 0.8336 & 0.5719 & 0.5492 \\
                           \cmidrule(lr){1-14}
$\alpha_{0}\beta_{0}$ (M)             & 0.5929 & 0.3219 & 0.6765 & 0.7356 & 0.5683 & 0.2790 & 0.4076 & 0.5819 & 0.6071 & 0.5687 & 0.8538 & 0.6310 & 0.5687 \\
$\alpha_{1}\beta_{0}$ (M)             & 0.5932 & 0.3198 & 0.6740 & 0.7364 & 0.5657 & 0.2835 & 0.4064 & 0.5815 & 0.6120 & 0.5667 & 0.8535 & 0.6309 & 0.5686 \\
\textbf{R. $\alpha_{1}\beta_{0}$* (M)} & 0.5876 & 0.3279 & 0.6773 & 0.7372 & 0.5674 & 0.2879 & 0.3770 & 0.5567 & 0.5496 & 0.5888 & 0.8315 & 0.6270 & 0.5597 \\
$\alpha_{1}\beta_{0.5}$ (M)           & 0.5863 & 0.3204 & 0.6626 & 0.7356 & 0.5683 & 0.2946 & 0.3929 & 0.5694 & 0.6304 & 0.5426 & 0.8514 & 0.6252 & 0.5650 \\
                           \midrule
All (L)                  & 0.6206 & 0.3562 & 0.7551 & 0.7167 & 0.5102 & 0.3192 & 0.3390 & 0.5030 & 0.3488 & 0.5187 & 0.8275 & 0.5646 & 0.5316 \\
\cmidrule(lr){1-14}
All (M)                  & 0.5796 & 0.3214 & 0.6257 & 0.7261 & 0.5623 & 0.2969 & 0.3782 & 0.5582 & 0.5985 & 0.5027 & 0.8569 & 0.6187 & 0.5521 \\
\bottomrule
\end{tabular}
\caption{Evaluation of LLaMA 3 8B Instruct (L) and Mistral 7B v0.3 Instruct (M) models on a variety of popular benchmarks to assess general reasoning, knowledge reasoning, and reading comprehension. The benchmarks used include MMLU0 (Massive Multitask Language Understanding Zero-shot), AGIEval (Advanced General Intelligence Evaluation), CSQA (CommonsenseQA), Winogrand (wg), ARC Challenge (ARC), GPQA0 (General-Purpose QA Zero-shot), TruthfulQA (TfQA) evaluated with multiple-choice (mc1, mc2) and generation BLEU scores, TriviaQA (TrQA), BoolQ, and HellaSwag (hs). The training configurations include Zero-shot (ZS), Fine-tuning (Ft.), and Retrieval-Augmented Fine-tuning (R.) with various $\alpha$ and $\beta$ settings for filtering data. The average scores indicate a slight performance drop for Mistral models compared to their respective baselines.}
\label{tab:lm_eval_other_bench}
\end{table*}

\section{Miscellaneous}
\label{misc}
Abbreviation Table can be found in Table \ref{tab:abbreviations}.

\begin{table}[htbp]
    \tiny
    \centering
    \begin{tabular}{ll}
    \toprule
    \textbf{Abbreviation} & \textbf{Full Form} \\
    \midrule
    BST   & Blended Skill Talk \\
    CA    & Conversational Agents \\
    Ft    & Fine-tuning \\
    ICC   & Intraclass Correlation Coefficient \\
    JIC   & Journal Intensive Conversation \\
    LLM   & Large Language Model \\
    OCEAN & Openness, Conscientiousness, Extraversion, Agreeableness, and Neuroticism \\
    P     & Pearson \\
    PAFt  & Persona Augmented Fine-tuning \\
    PC    & Persona Chat \\
    RAFt  & Retrieval-Augmented Fine-tuning \\
    RAG   & Retrieval-Augmented Generation \\
    Reg   & Regular \\
    S     & Spearman \\
    SPC   & Synthetic Persona Chat \\
    ZS    & Zero-shot \\
    \bottomrule
    \end{tabular}
    \caption{Abbreviation Table}
    \label{tab:abbreviations}
\end{table}

\end{document}